\documentclass[twoside]{article}

\usepackage{stfloats}

\usepackage{graphicx}
\usepackage{subfigure} 

\usepackage{url}

\usepackage{amssymb}
\usepackage{amsmath}

\usepackage[capitalize]{cleveref}

\usepackage{enumitem}
\usepackage{dsfont}
\usepackage{amsthm}
\usepackage{xfrac}

\newcommand{\norm}[1]{\lVert#1\rVert}
\newcommand{\Norm}[1]{\left\lVert#1\right\rVert}

\newcommand{\EE}[1]{\mathbb{E}\left[ #1 \right]}

\newcommand{\R}{\mathbb{R}}

\newcommand{\N}{\mathbb{N}}
\newcommand{\E}{\mathbb{E}}
\newcommand{\I}{\mathbb{I}}
\newcommand{\bP}{\mathbb{P}}

\newcommand{\calI}{\mathcal{I}}

\newcommand{\calB}{\mathcal{B}}
\newcommand{\calC}{\mathcal{C}}
\newcommand{\calD}{\mathcal{D}}

\newcommand{\calN}{\mathcal{N}}

\newcommand{\sa}{a}

\newcommand{\tdelta}{\tilde{\delta}}

\newcommand{\betahat}{\hat{\beta}}
\newcommand{\ud}{\mathrm{d}}

\newcommand{\Nlambda}{\lambda_{\scalebox{.6}{N}}}

\newcommand{\SigSS}{\tilde{\Sigma}_{SS}^{-1}}
\newcommand{\SigjS}{\tilde{\Sigma}_{jS}}
\newcommand{\Given}{\Big|}
\newcommand{\given}{\big|}

\newcommand{\iid}{\overset{iid}{\sim} }
\newcommand{\ie}{i.e.\ }

\newcommand{\st}{\mathrm{s.t.}}
\newcommand{\Unif}{\text{Unif}}
\newcommand{\jth}[1]{#1^{\mathrm{th}}}

\DeclareMathOperator*{\argmin}{\arg\!\min}

\DeclareMathOperator*{\supp}{supp}
\DeclareMathOperator*{\Var}{Var}

\newcounter{thmc}
\newcommand{\thmc}[1]{\refstepcounter{thmc}\label{#1}}
\newcounter{lmc}
\newcommand{\lmc}[1]{\refstepcounter{lmc}\label{#1}}
\newcounter{cc}

\newcounter{pc}
\newcommand{\pc}[1]{\refstepcounter{pc}\label{#1}}
\newcommand{\thm}[1]{{\bf Theorem \ref{#1}}}
\newcommand{\lemma}[1]{{\bf Lemma \ref{#1}}}
\newcommand{\prop}[1]{{\bf Proposition \ref{#1}}}

\newcommand{\thmfirst}[1]{\thmc{#1}\thm{#1}}
\newcommand{\lemmafirst}[1]{\lmc{#1}\lemma{#1}}

\newcommand{\propfirst}[1]{\pc{#1}\prop{#1}}

\usepackage[accepted]{aistats2014_arxiv}
\allowdisplaybreaks 
\begin{document}

%

%
\runningauthor{Oliva, P{\'o}czos , Verstynen, Singh, Schneider, Yeh, Tseng}

\twocolumn[

\aistatstitle{FuSSO: Functional Shrinkage and Selection Operator}

\aistatsauthor{ Junier B. Oliva$^\dagger$ \And Barnab{\'a}s P{\'o}czos$^\dagger$ \And Timothy Verstynen$^\dagger$ \And Aarti Singh$^\dagger$ }
\aistatsauthor{ Jeff Schneider$^\dagger$ \And Fang-Cheng Yeh$^\dagger$ \And Wen-Yih Tseng$^*$}

\aistatsaddress{ $^\dagger$Carnegie Mellon University \quad $^*$National Taiwan University } ]

\begin{abstract}
We present the FuSSO, a functional analogue to the LASSO, that efficiently finds a sparse set of functional input covariates to regress a real-valued response against. The FuSSO does so in a semi-parametric fashion, making no parametric assumptions about the nature of input functional covariates and assuming a linear form to the mapping of functional covariates to the response. We provide a statistical backing for use of the FuSSO via proof of asymptotic sparsistency under various conditions. Furthermore, we observe good results on both synthetic and real-world data.
\end{abstract}

\section{Introduction}

Modern data collection has allowed us to collect not just more data, but more complex data. In particular, complex objects like sets, distributions, and functions are becoming prevalent in many domains. It would be beneficial to perform machine learning tasks using these complex objects. However, many existing techniques can not handle complex, possibly infinite dimensional, objects; hence one often resorts to the ad-hoc technique of representing these complex object by arbitrary summary statistics. 

In this paper, we look to perform a regression task when dealing with functional data. Specifically, we look to regress a mapping that takes in many functional input covariates and outputs a real value. Moreover, since we are considering many functional covariates (possibly many more than the number of instances of one's data), we look to find an estimator that performs feature selection by only regressing on a subset of all possible input functional covariates. To this end we present the Functional Shrinkage and Selection Operator (FuSSO), for performing sparse functional regression in a principled, semi-parametric manner.

\begin{figure*}[t!]
        \centering
        \subfigure[Single Functional Covariate]{\includegraphics[width=.15\textwidth]{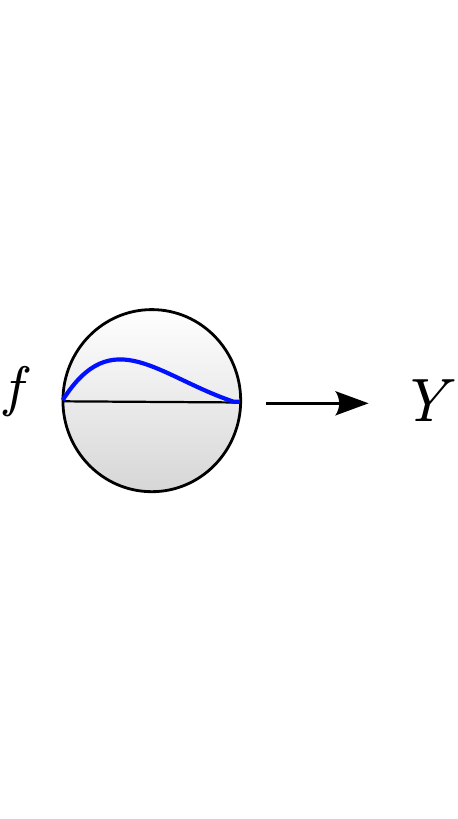}\label{fig:simp}}\quad\quad\quad
        \subfigure[Multiple Functional Covariates]{\includegraphics[width=.2\textwidth]{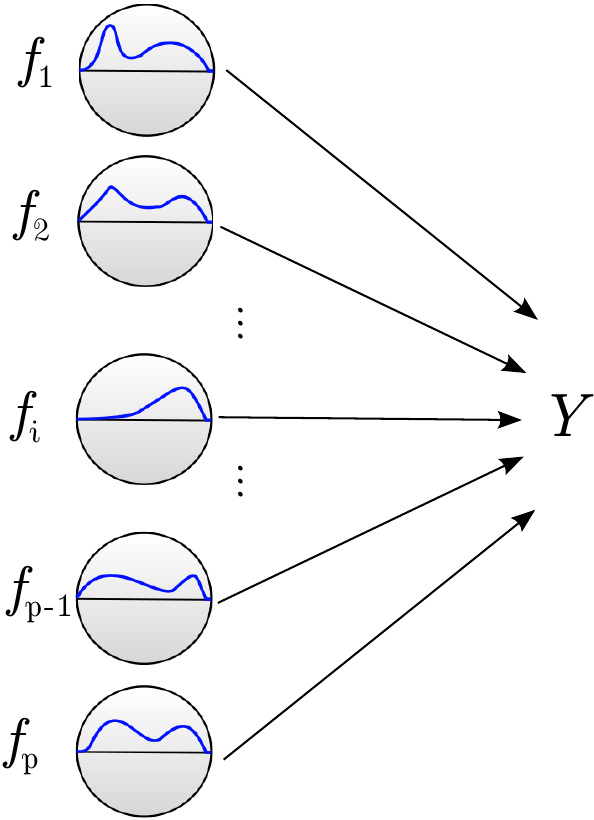}\label{fig:mult}}\quad\quad\quad
        \subfigure[Sparse Model]{\includegraphics[width=.2\textwidth]{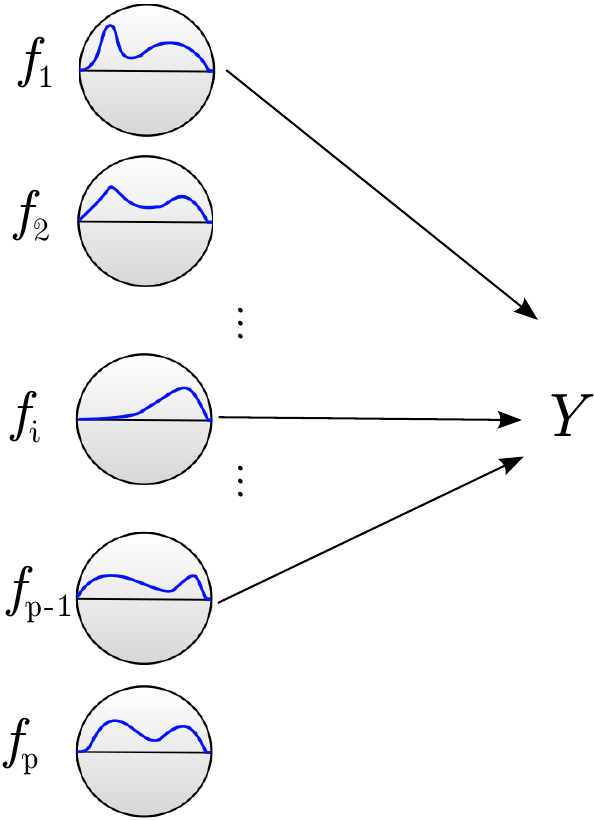}\label{fig:sparse}}
        \caption{(a) Model where mapping takes in a function $f$ and produces a real $Y$. (b) Model where response $Y$ is dependent on multiple input functions $f_{1},\ldots,f_{p}$. (c) Sparse model where response $Y$ is dependent on a sparse subset of input functions $f_{1},\ldots,f_{p}$.}
\vspace{-0.2cm}
\end{figure*}

Indeed, there are a multitude of applications and domains where the study of a mapping that takes in a functional input and outputs a real-value is of interest. That is, if $\calI$ is some class of input functions with domain $\Psi \subseteq \R$ and range $\R$, then one may be interested in a mapping $h:\calI\mapsto\R$: $h(f)=Y$ (Figure \ref{fig:simp}). Examples include: a mapping that takes in the time-series of a commodity's price in the past ($f$ is a function with the domain of time and range of price) and outputs the expected price of the commodity in the nearby future; also, a mapping that takes a patient's cardiac monitor's time-series and outputs a health index. Recently, work by \cite{poczos2012distfree} has explored this type of regression problem when the input function is a distribution. Furthermore, the general case of an arbitrary functional input is related to functional analysis \cite{ferraty2006nonparametric}.

However, often it is expected that the response one is interested in regressing is dependent on not just one, but many functions. That is, it may be fruitful to consider a mapping $h:\calI_1\times\ldots\times\calI_p\mapsto\R$: $h(f_1,\ldots,f_p)=Y$ (Figure \ref{fig:mult}). For instance, this is likely the case in regressing the price of a commodity in the future, since the commodity's future price is not only dependent on the history of it own price, but also the history of other commodities' prices as well. A response's dependence on multiple functional covariates is especially common in neurological data, where thousands of voxels in the brain may each contain a corresponding function. In fact, in such domains it is not uncommon to have a number of input functional covariates that far exceeds the number of training instances one has in a data-set. Thus, it would be beneficial to have an estimator that is sparse in the number of functional covariates used to regress the response against. That is, find an estimate, $\hat{h}_s$, that depends on a small subset $\{i_1,\ldots,i_{S}\} \subset \{1,\ldots,p\}$, such that $\hat{h}(f_1,\ldots,f_p)=\hat{h}_s(f_{i_1},\ldots,f_{i_{S}})$ (Figure \ref{fig:sparse}).  

Here we present a semi-parametric estimator to perform sparse regression with multiple input functional covariates and a real-valued response, the FuSSO: Functional Shrinkage and Selection Operator. No parametric assumptions are made on the nature of input functions. We shall assume that the response is the result of a sparse set of linear combinations of input functions and other non-paramteric functions $\{g_i\}$: $Y = \sum_j \langle f_j, g_j \rangle$. The resulting method is a LASSO-like \cite{tibshirani1996regression} estimator that effectively zeros out entire functions from consideration in regressing the response. 

Our contributions are as follows. We introduce the FuSSO, an estimator for performing regression with many functional covariates and a real-valued response. Furthermore, we provide a theoretical backing of the FuSSO estimator via proof of asymptotic sparsistency under certain conditions. We also illustrate the estimator with applications on synthetic data as well as in regressing the age of a subject when given orientation distribution function (dODF) \cite{yeh2011ntu} data for the subject's white matter.

\section{Related Work} 
As previously mentioned, recently \cite{poczos2012distfree} explored regression with a mapping that takes in a probability density function and outputs a real value. Furthermore, \cite{olivadistribution} studies the case when both the input and outputs are distributions. In addition, functional analysis relates to the study for functional data \cite{ferraty2006nonparametric}. In all these works, the mappings studied take in only one functional covariate. Based on them, it is not immediately evident how to expand on these ideas to develop an estimator that simultaneously performs regression and feature selection with multiple function covariates.

To the best of our knowledge, there has been no prior work in studying sparse mappings that take multiple functional inputs and produce a real-valued output. LASSO-like regression estimators that work with functional data include the following. In \cite{mingotti2013lasso}, one has a functional output and several real-valued covariates. Here, the estimator finds a sparse set of functions to scale by the real valued covariates to produce the functional response. Also, \cite{zhao2012wavelet,james2009functional} study the case when one has one functional covariate $f$ and one real valued response that is linearly dependent on $f$ and some function $g$: $Y = \langle f,g\rangle = \int f g$. In \cite{zhao2012wavelet} the estimator searches for sparsity across wavelet basis projection coefficients. In \cite{james2009functional}, sparsity is in achieved in the time (input) domain of the $d^{\mathrm{th}}$ derivative of $g$; i.e. $[D^dg](t)=0$ for many values of $t$ where $D^d$ is the differential operator. Hence, roughly speaking, \cite{zhao2012wavelet,james2009functional}  look for sparsity across frequency and time domains respectively, for the regressing function $g$. However, these methods do not consider the case where one has many input functional covariates $\{f_1,\ldots,f_p\}$, and needs to choose among them. That is, \cite{zhao2012wavelet,james2009functional} do not provide a method to select among function covariates in an analogous fashion to how the LASSO selects among real-valued covariates. 

Lastly, it is worth noting that in our estimator we will have an additive linear model, $\sum_j \langle f_j, g_j \rangle$ where we search for $\{g_i\}$ in a broad, non-parametric family such that many $g_j$ are the zero function. Such a task is similar in nature to the SpAM estimator \cite{ravikumar2009sparse}, in which one also has an additive model $\sum_j g_j(X_j)$ (in the dimensions of a real vector $X$) and searches for $\{g_i\}$ in a broad, non-parametric family such that many $g_j$ are the zero function. Note though, that in the SpAM model, the $\{g_i\}$ functions are applied to real covariates via a function evaluation. In the FuSSO model, $\{g_i\}$ are applied to functional covariates via an inner product; that is, FuSSO works over functional, not real-valued covariates, unlike SpAM.

\section{Model}
To better understand FuSSO's model we draw several analogies to real-valued linear regression and Group-LASSO \cite{yuan2006model}. Note that although for simplicity we focus on functions working over a one dimensional domain, it is straightforward to extend the estimator and results to the multidimensional case. Consider a model for typical real-valued linear regression with a data-set of input-output pairs $\{(X_i,Y_i) \}_{i=1}^{N}$:
\begin{align*}
Y_i = \langle X_i, w \rangle + \epsilon_i,
\end{align*}
where $Y_i \in \R$,$\ X_i \in \R^d$, $w\in\R^d,\ \epsilon_i \iid \calN(0,\sigma^2)$, and $\langle X_i, w \rangle=\sum_{j=1}^d X_{ij}w_{j}$. If instead one were working with functional data $\{(f^{(i)},Y_i)\}_{i=1}^{N}$, where $f^{(i)}:[0,1] \mapsto \R$ and $f^{(i)} \in L_2[0,1]$, one may similarly consider a linear model:
\begin{align*}
Y_i = \langle f^{(i)}, g \rangle + \epsilon_i,
\end{align*}
where, $g:[0,1] \mapsto \R$, and  $\langle f^{(i)}, g \rangle=\int_0^1 f^{(i)}(t)g(t) \ud t$. If $\Phi = \{\varphi_m\}_{m=1}^{\infty}$ is an orthonormal basis for $L_2[0,1]$ \cite{tsybakov2008introduction} then we have that 
\begin{align}
f^{(i)}(x)=\sum_{m=1}^{\infty}\alpha^{(i)}_{m}\varphi_m(x), \label{eq:projcoef}
\end{align}
where, $\alpha^{(i)}_{m} =\int_0^1 f^{(i)}(t)\varphi_m(t)\ud t$. Similarly, $g(x)=\sum_{m=1}^{\infty}\beta_{m}\varphi_m(x)$ where $\beta_{m} =\int_0^1 g(t)\varphi_m(t)\ud t$. Thus,
\begin{align*}
Y_i &= \langle f^{(i)}, g \rangle + \epsilon_i \\
&= \langle \sum_{m=1}^{\infty}\alpha^{(i)}_{m}\varphi_m(x), \sum_{k=1}^{\infty}\beta_{k}\varphi_k(x) \rangle + \epsilon_i \\
&= \sum_{m=1}^{\infty}\sum_{k=1}^{\infty} \alpha^{(i)}_{m}\beta_{k} \langle \varphi_m(x), \varphi_k(x) \rangle + \epsilon_i \\
 &= \sum_{m=1}^{\infty} \alpha^{(i)}_{m}\beta_{m} + \epsilon_i, 
\end{align*}
where the last step follows from orthonormality of $\Phi$.

Going back to the real-valued covariate case, if instead of having one feature vector per data instance: $X_i \in \R^d$, one had $p$ feature vectors associated to each data instance: $\{X_{ij}\ |\ 1\leq j \leq p,\ X_{ij} \in \R^d \}$, an additive linear model may be used for regression:
\begin{align*}
Y_i = \sum_{j=1}^p\langle X_{ij}, w_j \rangle + \epsilon_i, \mathrm{where}\ w_1,\ldots,w_p \in \R^d.
\end{align*}
Similarly, in the functional case one may have $p$ functions associated with data instance $i$: $\{f^{(i)}_{j}\ |\ \ 1\leq j \leq p,\ f^{(i)}_{j}\in L_2[0,1] \}$. Then, an additive linear model would be:
\begin{align}
Y_i &= \sum_{j=1}^p\langle f^{(i)}_j, g_j \rangle + \epsilon_i = \sum_{j=1}^p\sum_{m=1}^{\infty} \alpha^{(i)}_{jm}\beta_{jm} + \epsilon_i \label{eq:fgmodel},
\end{align}
where $g_1,\ldots,g_p \in L_2[0,1]$, and $\alpha^{(i)}_{jm}$ and $\beta_{jm}$ are projection coefficients for $f^{(i)}_j$ and $g_j$ respectively. 

Suppose that one has few observations relative to the number of features ($N\ll p$). In the real-valued case, in order to effectively find a solution for $ w = (w_1^{T},\ldots,w_p^{T})^T$ one may search for a group sparse solution where many $w_j=0$. To do so, one may consider the following Group-LASSO regression:
\begin{align}
w^\star = \argmin_{w} {\frac{1}{2N}\norm{Y-\sum_{j=1}^p X_j w_j}^2 + \Nlambda \sum_{j=1}^p \|w_j\| },
\label{eq:grplasso}
\end{align}
where $X_j$ is the $ N \times d $ matrix $X_j = [X_{1j} \ldots X_{Nj}]^T$, $Y=(Y_1,\ldots,Y_N)^T$, and $\|\cdot\|$ is the Euclidean norm.

If in the functional case (\ref{eq:fgmodel}) one also has that $N\ll p$, one may set up a similar optimization to (\ref{eq:grplasso}), whose direct analogue is: 
\begin{align}
g^\star = \argmin_{g}& \frac{1}{2N} \sum_{i=1}^N \left(Y_i - \sum_{j=1}^p \langle f^{(i)}_j, g_j \rangle \right)^2 \\
&+ \Nlambda \sum_{j=1}^p \norm{g_j} \label{eq:fgrplasso};
\end{align}
equivalently,
\begin{align}
\beta^\star = \argmin_{\beta}& \frac{1}{2N} \sum_{i=1}^N \left(Y_i - \sum_{j=1}^p\sum_{m=1}^{\infty} \alpha^{(i)}_{jm}\beta_{jm} \right)^2  \\
&+ \Nlambda \sum_{j=1}^p \sqrt{\sum_{m=1}^{\infty} \beta_{jm}^2}, \label{eq:bgrplasso} 
\end{align}
where $g = \{g_i\}_{i=1}^p =\{\sum_{m=1}^{\infty} \beta_{im}\varphi_m ,\}_{i=1}^p $.

However, it is unfeasible to directly observe functional inputs $\{f^{(i)}_j\ |\ 1\leq i \leq N, 1\leq j \leq p \}$ . Thus, we shall instead assume that one observes $\{\vec{y}^{\,(i)}_j\ |\ 1\leq i \leq N, 1\leq j \leq p \}$ where 
\begin{align}
\vec{y}^{\,(i)}_j =& \vec{f}^{\,(i)}_j + \xi^{(i)}_j, \label{eq:fvec}\\
\vec{f}^{\,(i)}_j =& \left(f^{(i)}_j(1/n),\ f^{(i)}_j(2/n),\ \ldots,\  f^{(i)}_j(1)\right)^T, \\
\xi^{(i)}_j \iid &\calN(0,\sigma_\xi^2I_n). 
\end{align} 
That is, we observe a grid of $n$ noisy values for each functional input. Then, one may estimate $\alpha^{(i)}_{jm}$ as:
\begin{align}
\tilde{\alpha}^{(i)}_{jm} = \frac{1}{n} \vec{\varphi}_m^T \vec{y}^{\,(i)}_j = \frac{1}{n} \vec{\varphi}_m^T  (\vec{f}^{\,(i)}_j + \xi^{(i)}_j ) = \bar{\alpha}^{(i)}_{jm} + \eta^{(i)}_{jm} \label{eq:talpha}
\end{align}
where $\vec{\varphi}_m = \left(\varphi_m(1/n),\ \varphi_m(2/n),\ \ldots,\  \varphi_m(1)\right)^T$. Furthermore, we may truncate the number of basis functions used to express ${f}^{\,(i)}_j$ to $M_n$, estimating it as:
\begin{align}
\tilde{f}^{(i)}_j(x) = \sum_{m=1}^{M_n} \tilde{\alpha}^{(i)}_{jm}  \varphi_m(x). \label{eq:trunc}
\end{align}
Using the truncated estimate (\ref{eq:trunc}), one has:
\begin{align*}
\langle \tilde{f}^{(i)}_j(x), g_j \rangle =& \sum_{m=1}^{M_n} \tilde{\alpha}^{(i)}_{jm} \beta_{jm}, \mathrm{and}\ \\
\norm{\tilde{f}^{(i)}_j(x)} =& \sqrt{\sum_{m=1}^{M_n} (\tilde{\alpha}^{(i)}_{jm})^2}.
\end{align*}
Hence, using the approximations (\ref{eq:trunc}), (\ref{eq:bgrplasso}) becomes:
\begin{align}
\betahat = \argmin_{\beta}& \frac{1}{2N} \sum_{i=1}^N \left(Y_i - \sum_{j=1}^p\sum_{m=1}^{M_n} \tilde{\alpha}^{(i)}_{jm}\beta_{jm} \right)^2  \label{eq:MSElasso}\\
&+ \Nlambda \sum_{j=1}^p \sqrt{\sum_{m=1}^{M_n} \beta_{jm}^2} \label{eq:normlasso}\\
= \argmin_{\beta}& \frac{1}{2N} \norm{Y - \sum_{j=1}^p\tilde{A}_{j}\beta_{j} }^2 + \Nlambda \sum_{j=1}^p \norm{\beta_{j}} \label{eq:btrunkgrplasso},
\end{align}
where $\tilde{A}_{j}$ is the $N \times M_n$ matrix with values $\tilde{A}_{j}(i,m) = \tilde{\alpha}^{(i)}_{jm}$ and $\beta_j = (\beta_{j1},\ldots,\beta_{jM_n})^T$. Note that one need not consider projection coefficients $\beta_{jm}$ for $m>M_n$ since such projection coefficients will not decrease the MSE term in \eqref{eq:MSElasso} (because $\tilde{\alpha}^{(i)}_{jm}=0$ for $m>M_n$), and $\beta_{jm}\neq0$ for $m>M_n$ increases the norm penalty term in (\ref{eq:normlasso}). Hence we see that our sparse functional estimates are a Group-LASSO problem on the projection coefficients. 

\paragraph{Extensions} It is useful to note that there are several straightforward extensions to the FuSSO as presented. First, we would like to note that it may be possible to estimate the inner product of a function $f^{(i)}_j$, and $g_j$ as $\int f^{(i)}_j g_j \approx \langle \vec{y}^{\,(i)}_j, \tfrac{1}{n}\vec{g}_j \rangle$, where $\vec{g}_j = (g_j(1/n),\ldots,g_j(1))^T$. This effectively allows one to use a naive approach of simply using Group-LASSO on the $\vec{y}^{\,(i)}_j$ feature vectors directly (we'll refer to this method as Y-GL). It is important to note, however, that Y-GL will be less robust to noise, and adaptive (and efficient) to smoothness than the FuSSO. Furthermore, we note that it is not necessary to have observations for input functions that are on a grid for the FuSSO, since one may estimate projection coefficients in the case of an irregular design \cite{tsybakov2008introduction}. Moreover, we may also estimate projection coeffincients for density functions with samples drawn from the pdf. Note that the Y-GL would fail to estimate our model in the irregular design case, and would not be possible in the case were functions are pdf. Also, a two-stage estimator as described in \cite{meinshausen2007relaxed}, where one first uses the regularization penalty with a large $\lambda$ to find the support, then solves the optimization problem with a smaller $\lambda$ on just the estimated support to estimate the response, may be more efficient at estimating the response. Furthermore, an analogous problem as (\ref{eq:btrunkgrplasso}) may be framed to perform logistic regression and classification. 

\section{Theory}
Next, we show that the FuSSO is able to recover the correct sparsity pattern asymptotically; i.e., that the FuSSO estimate is sparsistent. In order to do so, we shall show that with high probability there is an optimal solution to our optimization problem (\ref{eq:btrunkgrplasso}) with the correct sparsity pattern. We follow a similar argument to \cite{wainwright2006sharp, ravikumar2009sparse}. We shall use a ``witness" technique to show that there is a coefficient/subgradient pair $(\hat{\beta},\hat{u})$ such that $\supp(\hat{\beta})=\supp(\beta^*)$, for true response generating $\beta^{*}$. Let $\Omega(\beta) = \sum_{j=1}^{p} \norm{\beta_j}_2$, be our penalty term (\ref{eq:normlasso}). Let $S$ denote the true set of non-zero functions; \ie $S = \{ j\ |\ \beta^*_j\neq0 \}$, with $s = |S|$. First, we fix $\hat{\beta}_{S^c}=0$, and set $\hat{u}_{S} = \partial\Omega(\cdot)(\beta^*)_{S}$. Note that for a vector $\beta'$, $\partial\Omega(\cdot)(\beta') = \{ u \}$ where: $u_j = \beta'_j/\norm{\beta'}_2$, if $\beta'_j\neq0$; $u_j =\norm{u_j}_2 \leq 1$ if  $\beta'_j=0$.
 We shall show that with high probability, $\forall j \in S,\ \hat{\beta}_{j}\neq0$ and $\forall j \in S^c,\ \norm{u_j}_2 < 1$, thus showing that there is an optimal solution to our optimization problem (\ref{eq:btrunkgrplasso}) that has the true sparsity pattern with high probability.

First, we elaborate on our assumptions.

\subsection{Assumptions}
Let $\Phi$ be the trigonometric basis, $\varphi_1(x) \equiv 1$, $k\geq2$:
\begin{align*}
\varphi_{2k}(x) \equiv \sqrt{2}\cos(2\pi k x),\ 
\varphi_{2k+1}(x) \equiv \sqrt{2}\sin(2\pi k x)  .
\end{align*}
Let $\calD= \{ (\{\vec{y}^{\,(i)}_j\}_{j=1}^p, Y_i)\}_{i=1}^N$, where $\vec{y}^{\,(i)}_j$ is as (\ref{eq:fvec}), and $Y_i = \sum_{j=1}^p\sum_{m=1}^{\infty} \alpha^{(i)}_{jm}\beta^*_{jm} + \epsilon_i$ as in (\ref{eq:fgmodel}). Assume that $\forall\ 1\leq i \leq N,\ 1\leq j \leq p$: $\alpha^{(i)}_j\in \Theta(\gamma,Q)$, where:
\begin{align*}
\Theta(\gamma,Q) =& \{ \theta\ :\ \sum_{k=1}^\infty c_k^{2} \theta_k^2 \leq Q \},\\
c_k =& k^\gamma\  \mathrm{if}\ k\ \mathrm{even\ or\ one},\ (k-1)^\gamma\  \mathrm{otherwise},\\ 
\alpha^{(i)}_j =& \{\alpha^{(i)}_{jm} \in \R\ |\ \alpha^{(i)}_{jm} = \int_0^1 f^{(i)}_j \varphi_m,\ m\in\N^{+} \}
\end{align*} 
for $0<Q<\infty$ and $\tfrac{1}{2}<\gamma<\infty$. Furthermore, assume that that for the true $\beta$ generating the observed responses $Y_i$, $\beta^{*}$, $\forall\ 1\leq j \leq p$: $\beta^{*}_j \in \Theta(\gamma,Q)$.

Let $A_j$ be the $N \times M_n$ matrix with entries $A_{j}(i,m) = \alpha^{(i)}_{jm}$. Let $A_S$ denote the matrix made up from horizontally concatenating the $A_j$ matrices with $j\in S$; \ie $A_S = [A_{j_1}\ldots A_{j_s}]$, where $\{j_1,\ldots,j_s\}=S$ and $j_i<j_k$ for $i<k$.  Suppose the following:
\begin{align}
\Lambda_{\max} \left(\tfrac{1}{N}A_S^TA_S\right) \leq C_{\max} <& \infty \label{eq:maxeig} \\
\Lambda_{\min} \left(\tfrac{1}{N}A_S^TA_S\right) \geq C_{\min} >& 0 \label{eq:mineig}.
\end{align}
Also, suppose $\exists\delta\in(0,1]$ s.t. $\forall j\in S^c$
\begin{align}
\Lambda_{\max} \left(\tfrac{1}{N}A_j^TA_j\right) \leq C_{\max} <& \infty  \\
\Norm{(\tfrac{1}{N}A_{j}^TA_S)(\tfrac{1}{N}A_S^TA_S)^{-1}}_2 \leq&  {1-\delta}/{\sqrt{s}}
\end{align}
Let $\bar{A}_j$ be the $N \times M_n$ matrix with entries $\bar{A}_j(i,m) =  \bar{\alpha}^{(i)}_{jm} = \frac{1}{n} \vec{\varphi}_m^T  \vec{f}^{\,(i)}_j  $. Let $H_j$ be the $N \times M_n$ matrix with entries $H_{j}(i,m) = \eta^{(i)}_{jm} = \frac{1}{n} \vec{\varphi}_m^T  \xi^{(i)}_j $. Thus, $\tilde{A}_j = \bar{A}_j + H_j$. Furthermore, let $E_j = \bar{A}_j - A_j$. Then, $\tilde{A}_j =A_j +E_j+ H_j$.

In addition to the aforementioned assumptions, we shall further assume the following:
\begin{gather}
\exists \sa<1/2 \quad \st\quad pM_n n^{\sa - \sfrac{1}{2}} e^{-n^{1-2\sa}} \rightarrow 0 \label{eq:bndH_S}\\
\rho^*_N \equiv \min_{j\in S}\norm{\beta^*_j}_\infty>0 \label{eq:supportaway}\\
\sqrt{s M_n} \left(n^{-\gamma +\sfrac{1}{2}}+ n^{-\sa} \right) \rightarrow 0 \\
\frac{1}{\rho^*_N}\left( s^{\sfrac{3}{2}}M_n^{\sfrac{1}{2}-2\gamma} + \sqrt{{\log(sM_n)}/{N}} \right) \rightarrow 0 \\
\Nlambda \sqrt{sM_n}/\rho^*_N \rightarrow 0 \label{eq:lambdaroot}\\
\frac{1}{\Nlambda}\left({s\sqrt{M_n}} n^{-\gamma +1/2} +   \sqrt{\tfrac{s\log(N)}{n}} \right) \rightarrow 0 \\
\frac{1}{\Nlambda} \left(  \frac{s M_n}{ n^{\gamma+\sa -\sfrac{1}{2}}}   + \frac{\sqrt{sM_n\log(N)}}{ n^{\sa +\sfrac{1}{2}}} \right) \rightarrow 0\\
\frac{1}{\Nlambda} \sqrt{{M_n\log((p-s)M_n)}/{N}}  \rightarrow 0 \\
{s}/{(\Nlambda N M_n^{2\gamma-\sfrac{1}{2}})} \rightarrow 0 \label{eq:l-ass},
\end{gather}
and we assume $\gamma \geq 1$ for the sake of simplification. We may further simplify our assumptions if we take $n=N^{1/2}$ and choose $M_n$ optimally for function estimation: $M_n\asymp n^{1/(2\gamma +1)}=N^{1/(4\gamma +2)}$. Furthermore, take $s = O(1)$, $\rho^*_N \asymp 1$, and $\gamma=2$. Under these conditions, our assumptions reduce to $\frac{1}{10}<\sa$ and taking the follow to go to zero:
\begin{gather*}
pN^{\frac{10a-3}{20}}e^{-N^{\frac{1}{2}-a}},
\Nlambda N^{\sfrac{1}{20}}, N^{-\frac{7}{10}}/\Nlambda,\\
\tfrac{1}{\Nlambda^2} N^{\frac{1}{2}}\log(N),\tfrac{1}{\Nlambda^2} N^{-\sfrac{9}{10}}\log(pN) .
\end{gather*}

\subsection{Sparsistency}

\thmfirst{thm:sparsistent}: $\bP\left(\hat{S}_N=S\right) \rightarrow 1$.

First, we state some lemmas, whose proofs may be found in the supplementary materials.
\subsubsection{Lemmata}

\lemmafirst{thm:exp_bnd} Let $X$ be a non-negative r.v. and $\calC$ be an measurable event, then $\EE{X | \calC }\bP(\calC) \leq \EE{X} $. 

\lemmafirst{thm:sumkrondelt} $\frac{1}{n}\sum_{k=1}^n \varphi_{m}(k/n)\varphi_{l}(k/n) = \I\{l=m\}$, for $1\leq l,m \leq n-1$. 

\lemmafirst{thm:noisedist} Let $H_{j}^{(i)}$ be the rows of $H_{j}$, then $H_{j}^{(i)} \iid \calN(0,\frac{\sigma_\xi^2}{n}I)$, and $H_{S}^{(i)} \iid \calN(0,\frac{\sigma_\xi^2}{n}I)$.


\lemmafirst{thm:bndHmax} $\bP\left(\norm{H}_{\max} \geq n^\sa \right) \leq 2 \sigma_\xi pM_n n^{\sa - \sfrac{1}{2}} e^{-\frac{n^{1-2\sa}}{2\sigma_\xi^2}}$

\lemmafirst{thm:biasmax} $\norm{E_j}_{\max}\leq C_Q n^{-\gamma+1/2}$, where $C_Q\in(0,\infty)$ is a constant depending on $Q$. 

\lemmafirst{thm:beta} $\norm{\beta^{*}_S}_2^2 \leq Qs$. 

\lemmafirst{thm:noiseeigs} $\exists N_0,n_0,\tilde{C}_{\min}, \tilde{C}_{\max}$, $ 0<\tilde{C}_{\min}\leq \tilde{C}_{\max} < \infty$, $0<\tdelta\leq1$ s.t. if $\norm{H}_{\max}<n^{-a}$, and $N>N_0$, $n>n_0$ then
\begin{align}
\Lambda_{\max} \left(\tfrac{1}{N}\tilde{A}_S^T\tilde{A}_S\right) \leq \tilde{C}_{\max} <& \infty \\
\Lambda_{\min} \left(\tfrac{1}{N}\tilde{A}_S^T\tilde{A}_S\right) \geq \tilde{C}_{\min} >& 0\\
\forall j \in S^c,\ \Norm{(\tfrac{1}{N}\tilde{A}_{j}^T\tilde{A}_S)(\tfrac{1}{N}\tilde{A}_S^T\tilde{A}_S)^{-1}}_2 \leq&  \frac{1-\tdelta}{\sqrt{s}}
\end{align}

\subsubsection{Proof of Theorem 1}




\propfirst{thm:infsupp} $\bP\left(\forall j\in S\ \hat{\beta}_j\neq0\right) \rightarrow 1$.
\begin{proof}
Recall that by (\ref{eq:supportaway}), $\rho^*_N = \min_{j\in S}\norm{\beta^*_j}_\infty>0$. Thus to prove that $\forall j\in S\ \hat{\beta}_j\neq0$,  it suffices to show that :
$\norm{\hat{\beta}_S-\beta^{*}_S}_\infty \leq \frac{\rho^*_N}{2}$.
To do so we show $\bP\left(\norm{\hat{\beta}_S-\beta^{*}_S}_\infty > \frac{\rho^*_N}{2}\right) \rightarrow 0$. Let $\calB$ be the event that $\norm{H}_{\max}<n^{-\sa}$. Note that:
\begin{align*}
&\bP\left(\norm{\hat{\beta}_S-\beta^{*}_S}_\infty > \frac{\rho^*_N}{2}\right)\\
&\leq \bP\left(\norm{\hat{\beta}_S-\beta^{*}_S}_\infty > \frac{\rho^*_N}{2} \Given \calB \right) \bP(\calB) +\bP\left(\calB^c \right).
\end{align*}
Furthermore,
\begin{align*}
&\bP\left(\norm{\hat{\beta}_S-\beta^{*}_S}_\infty > \frac{\rho^*_N}{2} \Given \calB \right) \leq \tfrac{2}{\rho^*_N}\E\left[\norm{\hat{\beta}_S-\beta^{*}_S}_\infty \Given \calB\right].
\end{align*}

Then, looking at the stationarity condition for the support $S$:
\begin{align}
\tfrac{1}{N}\tilde{A}_S^T\left(\tilde{A}_S\hat{\beta}_S-Y\right)+\Nlambda\hat{u}_S=0. \label{eq:supp-station}
\end{align}
Let $V$ be the $N \times 1$ vector with entries $V_i= \sum_{j\in S}\sum_{m=M_n+1}^{\infty}\alpha^{(i)}_{jm}\beta^{*}_{jm}$; \ie the error from truncation. Then, using \eqref{eq:supp-station} $Y=A_S\beta^{*}_S +V+\epsilon \implies$
\begin{align*}
&\tfrac{1}{N}\tilde{A}_S^T\left(\tilde{A}_S\hat{\beta}_S-A_S\beta^{*}_S -V-\epsilon\right)+\Nlambda\hat{u}_S=0  \implies \\
&\tfrac{\tilde{A}_S^T}{N}\left(\tilde{A}_S(\hat{\beta}_S-\beta^{*}_S)-(A_S-\tilde{A}_S)\beta^{*}_S -V-\epsilon\right)=
-\Nlambda\hat{u}_S
\end{align*}

Thus,
\begin{align}
\tfrac{1}{N}\tilde{A}_S^T\tilde{A}_S(\hat{\beta}_S-\beta^{*}_S)= &-\tfrac{1}{N}\tilde{A}_S^T(E_S+H_S)\beta^{*}_S +\tfrac{1}{N}\tilde{A}_S^TV \nonumber \\
&+\tfrac{1}{N}\tilde{A}_S^T\epsilon-\Nlambda\hat{u}_S \label{eq:supp-diff}.
\end{align}
Let $\SigSS=(\tfrac{1}{N}\tilde{A}_S^T\tilde{A}_S)^{-1}$; we see that,
\begin{align*}
\norm{\hat{\beta}_S-\beta^{*}_S}_\infty & \leq  \norm{\SigSS(\tfrac{1}{N}\tilde{A}_S^T)(E_S+H_S)\beta^{*}_S}_\infty \\
& +\norm{\SigSS(\tfrac{1}{N}\tilde{A}_S^T)V}_\infty +\norm{\SigSS(\tfrac{1}{N}\tilde{A}_S^T)\epsilon}_\infty\\
& +\norm{\SigSS\Nlambda\hat{u}_S}_\infty.
\end{align*}

Thus, we proceed to bound each term on the LHS in expectation. First, note that $\norm{\SigSS(\tfrac{1}{N}\tilde{A}_S^T)(E_S+H_S)\beta^{*}_S}_\infty 
\leq \norm{\SigSS}_\infty\norm{(\tfrac{1}{N}\tilde{A}_S^T)(E_S+H_S)\beta^{*}_S}_\infty 
= \norm{\SigSS}_\infty \norm{\tfrac{1}{N}(A_S^T +E_S^T+ H_S^T)(E_S+H_S)\beta^{*}_S}_\infty  
\leq  \frac{\norm{\SigSS}_\infty}{N} \Big( \norm{A_S^T(E_S+H_S)\beta^{*}_S}_\infty
+ \norm{(E_S+H_S)^T(E_S+H_S)\beta^{*}_S}_\infty \Big)$.
Moreover, given that $\calB$ occurs:
\begin{align*}
\norm{\SigSS}_\infty \leq \sqrt{sM_n} \norm{\SigSS}_2 \leq \frac{\sqrt{sM_n} }{\tilde{C}_{\min}}.
\end{align*}
Thus, $\E\left[ \tfrac{\norm{\SigSS}_\infty}{N} \norm{A_S^T(E_S+H_S)\beta^{*}_S}_\infty \given \calB \right]\bP(\calB)$
\begin{align*}
&\leq \frac{\sqrt{sM_n} }{\tilde{C}_{\min}N} \norm{A_S^T}_\infty \E\left[ \norm{(E_S+H_S)\beta^{*}_S}_\infty \given \calB \right] \bP(\calB)\\
&\leq \frac{Q\sqrt{sM_n} }{\tilde{C}_{\min}} \left(\norm{E_S\beta^{*}_S}_\infty + \E\left[ \norm{H_S\beta^{*}_S}_\infty \given \calB \right]\bP(\calB) \right),
\end{align*}
noting that $\norm{A_S^T}_{\infty}\leq NQ$. Moreover, by Lemma \ref{thm:exp_bnd}:
\begin{align*}
\E\left[ \norm{H_S\beta^{*}_S}_\infty \given \calB \right]\bP(\calB) \leq \E\left[ \norm{H_S\beta^{*}_S}_\infty \right].
\end{align*}
Also, $H_S\beta^{*}_S$ is normally distributed and $\Var[H^{(i)T}_S\beta^{*}_S]$
\begin{align*}
 = \sum_{j=1}^{sM_n} \Var[H^{(i)}_{Sj}\beta^{*}_{Sj}] =  \frac{\sigma_\xi^2}{n} \norm{\beta^{*}_{S}}_2^2 
\leq  \frac{\sigma_\xi^2Q s}{n} .
\end{align*}
Hence, by a Gaussian inequality (e.g. \cite{wasserman10705}) we have:
\begin{align*}
\E\left[ \norm{H_S\beta^{*}_S}_\infty \right] \leq \sqrt{{2\sigma_\xi^2Q s \log(N)}/{n} }.
\end{align*}
Unless otherwise specified, let $X^{(i)}$ be the $\jth{i}$ row of matrix $X$ and $X_j$ be the $\jth{j}$ column. Also,
\begin{align*}
\norm{E_S\beta^{*}_S}_\infty &= \max_{1\leq i \leq N} |E_S^{(i)T}\beta^{*}_S| \leq \norm{\beta^{*}_S}_2\max_{1\leq i \leq N} \norm{E_S^{(i)}}_2 \\
& \leq\sqrt{Qs}\left( C_Q \sqrt{sM_n} n^{-\gamma +1/2} \right) \\
& = \sqrt{Q}C_Q s\sqrt{M_n} n^{-\gamma +1/2} 
\end{align*}
Thus, $\E\left[ \tfrac{\norm{\Sigma_{SS}^{-1}}_\infty}{N} \norm{A_S^T(E_S+H_S)\beta^{*}_S}_\infty \given \calB \right]$
\begin{align*}
= O\left( \sqrt{sM_n} \left( s\sqrt{M_n} n^{-\gamma +1/2} +  \sqrt{\frac{s\log(N)}{n} } \right) \right).
\end{align*}
Furthermore, $\E\left[ \norm{(E_S+H_S)^T(E_S+H_S)\beta^{*}_S}_\infty \given \calB \right]\bP(\calB)$
\begin{align*}
& =\E\left[ \max_{j \leq sM_n} |(E_{Sj}+H_{Sj})^T\left((E_{S}+H_{S})\beta_S^*\right)| \given \calB \right]\bP(\calB) \\
& \leq \E\left[ \max_{j\leq sM_n}  \norm{E_{Sj}+H_{Sj}}_1 \norm{(E_{S}+H_{S})\beta_S^*}_\infty \given \calB \right]\bP(\calB) \\
& = \E\left[ \norm{E_{S}+H_{S}}_{1} \norm{(E_{S}+H_{S})\beta_S^*}_\infty \given \calB \right]\bP(\calB).
\end{align*}
Then, given that $\calB$ occurs $\norm{E_{S}+H_{S}}_{1}$
\begin{align*}
 \leq \norm{E_{S}}_{1}+\norm{H_{Sj}}_{1} 
\leq  N(C_Q n^{-\gamma+1/2}+n^{-\sa}),
\end{align*}
and, as before: $\EE{\norm{(E_{S}+H_{S})\beta_S^*}_\infty | \calB}$
\begin{align*}
&\leq C_2 s\sqrt{M_n} n^{-\gamma +1/2} + C_3 \sqrt{\frac{s\log(N)}{n} }.
\end{align*}
Hence, $\EE{\norm{\SigSS(\tfrac{1}{N}\tilde{A}_S^T)(E_S+H_S)\beta^{*}_S}_\infty \Given \cal B}\bP(\calB)$
\begin{align*}
=& O\Big(\sqrt{sM_n}\left(  s\sqrt{M_n} n^{-\gamma +1/2} + \sqrt{{s\log(N)}/{n} } \right)\\
&\left(1+ n^{-\gamma+1/2}+n^{-\sa}\right)\Big)\\
=& O\left( \sqrt{sM_n} \left( s\sqrt{M_n} n^{-\gamma +1/2} +  \sqrt{{s\log(N)}/{n} } \right)\right)
\end{align*}

The next terms are bounded as follows\footnote{See Supplemental Materials for proof.}.

\lemmafirst{thm:Vterm} $\E\left[\norm{\SigSS(\tfrac{1}{N}\tilde{A}_S^T)V}_\infty \Given \calB \right]\bP(\calB) = O\left( \frac{s^{\sfrac{3}{2}}}{M_n^{2\gamma-\sfrac{1}{2}}} \right)$ .

\lemmafirst{thm:epsterm}  $\EE{\norm{\SigSS(\tfrac{1}{N}\tilde{A}_S^T)\epsilon}_\infty \Given \calB}\bP(\calB)  = O\left(\sqrt{\log(sM_n)/N}\right)$.

Lastly,
\begin{align*}
&\norm{\hat{u}_S}_\infty = \max_{j\in S} \norm{\hat{u}_j}_\infty \leq \max_{j\in S} \norm{\hat{u}_j}_2 \leq 1 \implies\\
&\norm{\SigSS\Nlambda\hat{u}_S}_\infty \leq \Nlambda\norm{\SigSS}_\infty \norm{\hat{u}_S}_\infty \leq \frac{\Nlambda\sqrt{sM_n} }{\tilde{C}_{\min}}.
\end{align*}
Keeping only leading terms, $\EE{\norm{\hat{\beta}_S-\beta^{*}_S}_\infty \Given \calB}\bP(\calB) $
\begin{align*}
=&O\left( s^{\sfrac{3}{2}} M_n n^{-\gamma +1/2} + s\sqrt{{M_n\log(N)}/{n}} \right)\\
&+O\left({s^{\sfrac{3}{2}}}/{M_n^{2\gamma-\sfrac{1}{2}}}+\sqrt{{\log(sM_n)}/{N}} + \Nlambda\sqrt{sM_n}\right).
\end{align*}
Hence, by assumptions (\ref{eq:bndH_S})-(\ref{eq:lambdaroot}) we have $\bP\left(\norm{\hat{\beta}_S-\beta^{*}_S}_\infty > \frac{\rho^*_N}{2}\right) \rightarrow 0$
\end{proof}


One may similarly look at the stationarity for $j\in S^c$ to analyze $\hat{u}_j$: $0=\tfrac{1}{N}\tilde{A}_j^T\left(\tilde{A}_S\beta_S-Y\right)+\Nlambda\hat{u}_j$
\begin{align*}
=&\tfrac{\tilde{A}_j^T}{N}\left(\tilde{A}_S(\hat{\beta}_S-\beta^{*}_S)-(A_S-\tilde{A}_S)\beta^{*}_S -V-\epsilon\right) +\Nlambda\hat{u}_j.
\end{align*}
Thus,
\begin{align*}
\hat{u}_j =&\tfrac{1}{\Nlambda N}\tilde{A}_j^T\tilde{A}_S(\beta^{*}_S-\hat{\beta}_S)+\tfrac{1}{\Nlambda N}\tilde{A}_j^T(A_S-\tilde{A}_S)\beta^{*}_S \\
&+ \tfrac{1}{\Nlambda N} \tilde{A}_j^T (V+\epsilon)\\
=&\tfrac{1}{\Nlambda}\SigjS\SigSS\Big( \tfrac{1}{N}\tilde{A}_S^T(E_S+H_S)\beta^{*}_S -\tfrac{1}{N}\tilde{A}_S^TV\\ &-\tfrac{1}{N}\tilde{A}_S^T\epsilon+\Nlambda\hat{u}_S\Big)-\tfrac{1}{\Nlambda  N}\tilde{A}_j^T(E_S+H_S)\beta^{*}_S \\
&+ \tfrac{1}{\Nlambda N} \tilde{A}_j^T (V+\epsilon),
\end{align*}
where $\SigjS=\tfrac{1}{N}\tilde{A}_j^T\tilde{A}_S$ and using \eqref{eq:supp-diff}. We wish to show that $\forall j\in S^c$ $\hat{u}_j$ satisfies the KKT conditions, that is:

\propfirst{thm:muH} $\bP\left(\max_{j\in S^c} \norm{\hat{u}_j}_2 < 1 \right) \rightarrow 1$.
\begin{proof}
Let $\mu_j^H \equiv \EE{\hat{u}_j \given H}$. We proceed as follows:
\begin{align*}
&\bP\left(\max_{j\in S^c} \norm{u_j}_2 < 1 \right) \\
&\ge  \bP\left(\max_{j\in S^c} \norm{\mu_j^H}_2 + \norm{u_j - \mu_j^H}_2 < 1 \right)\\
&\ge  \bP\left(\max_{j\in S^c} \norm{\mu_j^H}_2 + \sqrt{M_n} \norm{u_j - \mu_j^H}_\infty < 1 \right) \\
&\ge \bP \left(\max_{j\in S^c} \norm{\mu_j^H}_2< 1-\tfrac{\tdelta}{2},\max_{j\in S^c}  \norm{u_j - \mu_j^H}_\infty < \tfrac{\tdelta}{2\sqrt{M_n}} \right) \\
&\ge 1 - \bP\left(\max_{j\in S^c} \norm{\mu_j^H}_2 \geq 1-\tdelta \right) \\
&- \bP\left(\max_{j\in S^c} \norm{u_j - \mu_j^H}_\infty \geq \frac{\tdelta}{2\sqrt{M_n}} \right).
\end{align*}

We obtain the following results:

\lemmafirst{thm:uH} $\bP\left(\max_{j\in S^c} \norm{\mu_j^H }_2 \geq  1-\frac{\tdelta}{2} \right) \rightarrow 0 $

\lemmafirst{thm:uj} $\bP\left(\max_{j\in S^c} \norm{\hat{u}_j - \mu_j^H }_\infty \geq  \frac{\tdelta}{2\sqrt{M_n}} \right) \rightarrow 0$

Hence, we have that $\bP\left(\max_{j\in S^c} \norm{\hat{u}_j}_2 < 1 \right) \rightarrow 1$. 
\end{proof}

\section{Experiments}

\subsection{Synthetic Data}
\begin{figure}[t]
        \centering
        \subfigure[Function at $n=5$]{\includegraphics[width=.23\textwidth]{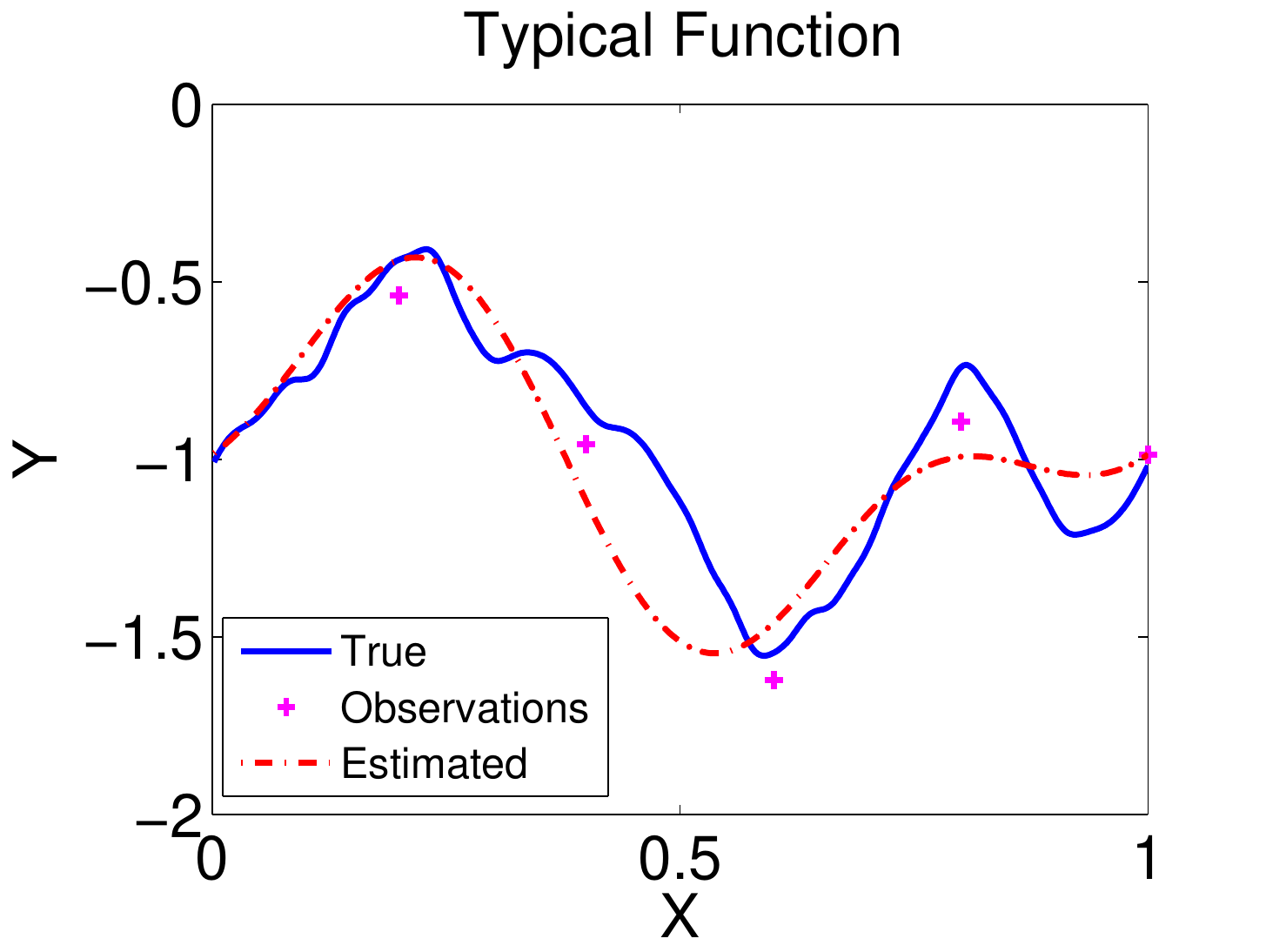}}
        \subfigure[Reg. Path at $p=100$, $n=50$, $n=5$ of $\norm{\hat{\beta}_j}$]{\includegraphics[width=.23\textwidth]{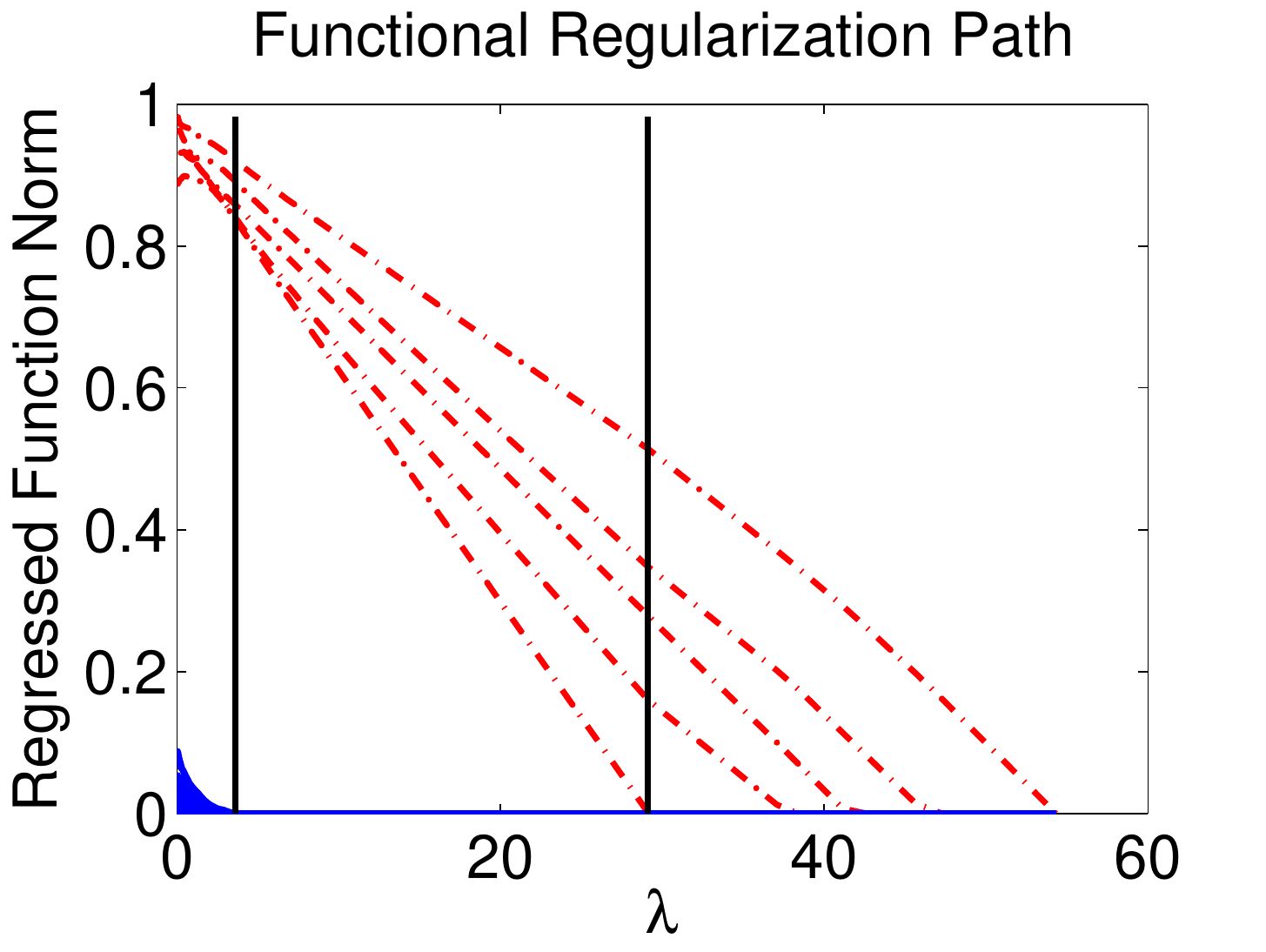}\label{fig:func_n5}}
        \subfigure[Function at $n=25$]{\includegraphics[width=.23\textwidth]{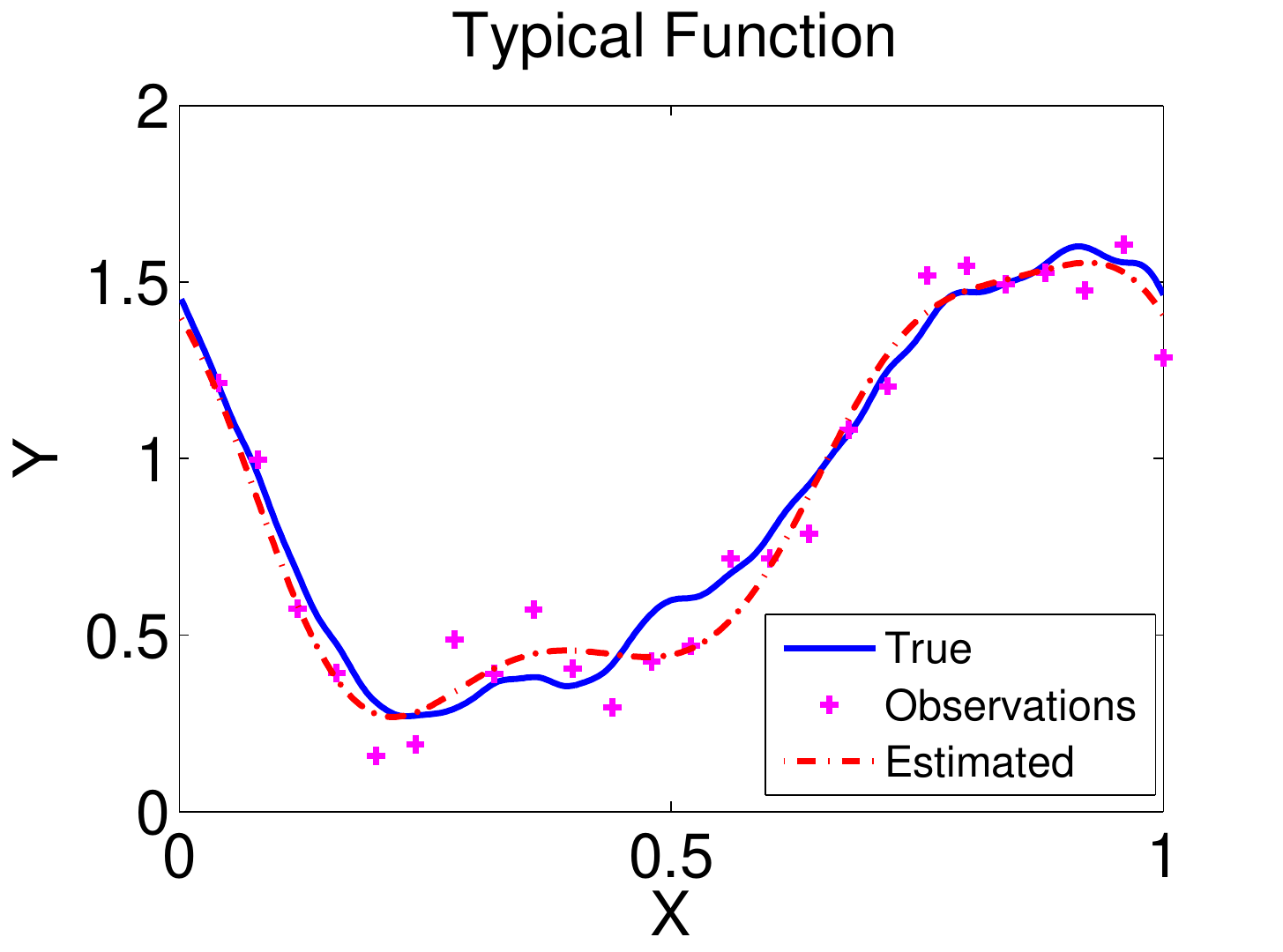}\label{fig:func_n25}}
        \subfigure[Reg. Path at $p=1000$, $n=500$, $n=25$ of $\norm{\hat{\beta}_j}$]{\includegraphics[width=.23\textwidth]{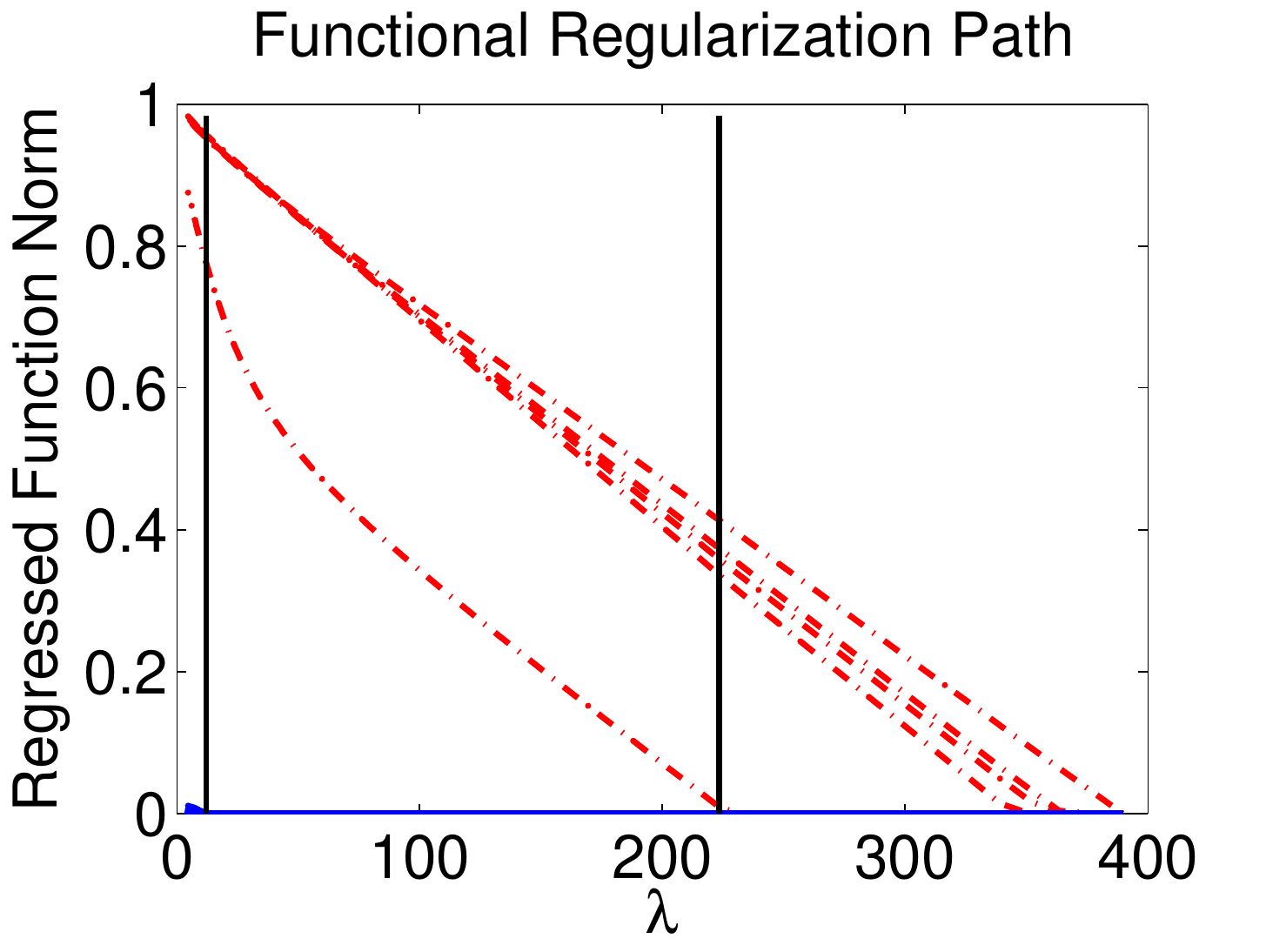}}
        \caption{(a)(c) Two typical functions, noisy observations, and estimates. (b)(d) Regularization paths showing the norms of $\hat{\beta}_j$ (in red for $j$ in support, blue otherwise) for a range of $\lambda$; rightmost vertical line indicates largest $\lambda$ able to recover the support, leftmost line for smallest such $\lambda$}
\vspace{-0.3cm}
\end{figure}

We tested the FuSSO on synthetic data-sets of $\calD= \{ (\{\vec{y}^{\,(i)}_j\}_{j=1}^p,Y_i)\}_{i=1}^N$ (where $\vec{y}^{\,(i)}_j$ as in (\ref{eq:fvec})). The experiments performed were as follows. First, we fix $N,n,p,$ and $s$. For $i=1,\ldots,N$, $j=1,\ldots,p$ we create random functions using a maximum of $M$ projection coefficients as follows: 1) Set $a_{jm}\iid\Unif[-1,1]$ for $m=1,\ldots,M$; 2) set $a_{jm}= a_{ji}/c_m^2$, where $c_m = m$ if $m=1$ or is even, $c_m = m-1$ if $m$ is odd; 3) set $a_{jm} = a_{jm}/\norm{a_j}$; 4) set $\alpha^{(i)}_j = a_j$. (See Figures \ref{fig:func_n5},\ref{fig:func_n25} for typical functions.) Similarly, we generate $\beta_j^*$ for $j=1,\ldots,s$; for $j=s+1,\ldots,p$, we set $\beta^*_j = 0$. Then, we generate $Y_i$ as $Y_i = \sum_{j=1}^p \langle\beta^*_j,\alpha^{(i)}_j\rangle + \epsilon_i = \sum_{j=1}^s \langle\beta^*_j,\alpha^{(i)}_j\rangle + \epsilon_i $, where $\epsilon\iid\calN(0,.1)$. Also, a grid of $n$ noisy function evaluations were generated to make $\vec{y}^{\,(i)}_j$ as in (\ref{eq:fvec}), with $\sigma_\xi=.1$. These were then used to compute $\tilde{\alpha}^{(i)}_{jm}$ for $m=1,\ldots,M_n$ as in (\ref{eq:talpha}), $M_n$ was chosen by cross validation. (See Figures \ref{fig:func_n5}, \ref{fig:func_n25} for typical noisy observations and function estimates for $n=5$ and $n=25$ respectively.)

We fixed $s=5$ and chose the following configurations for the other parameters: $(p,N,n)\in\{(100,50,5),(1000,500,25),(20000,500,25)\}$. For each tuple of $(p,N,n)$ configurations, 100 random trails were performed. We recorded, $r$, the fraction of the trails that a $\lambda$ value was able recover the correct sparsity pattern (\ie that only the first 5 functions are in the support). We also recorded the mean length of the range of $\lambda$, $\Delta_\lambda$, that were able to recover the correct support; \ie $\Delta_\lambda=\tfrac{1}{t}\sum_{t=1}^{100}\Delta^{(t)}_\lambda$, where $\Delta^{(t)}_\lambda= (\lambda^{(t)}_f-\lambda^{(t)}_l)/\lambda^{(t)}_{\max}$, $\lambda^{(t)}_f$ is the largest $
\lambda$ value found to recover the correct support in the $\jth{t}$ trails, $\lambda^{(t)}_l$ the smallest such $\lambda$, and $\lambda^{(t)}_{\max}$ is the smallest $\lambda$ to produce $\hat{\beta}=0$ ($\Delta^{(t)}_\lambda$ is taken to be zero if no $\lambda$ recovered the correct support). The results were as follows: 
\begin{center}
\begin{tabular}{l{c}{c}}
$(p,N,n)$         & $r$ & $\Delta$  \\
\hline
(100,50,5) 				 & .68 & .2125   \\
(1000,500,25)            & 1   & .4771   \\
(20000,500,25)           & 1   & .4729  \\
\end{tabular}
\end{center}
Hence we see that even when the number of observations per function is small ($5$ or $25$) and the number of total number of input functional covariates is large (we were able to test up to $20000$), the FuSSO can recover the correct support.
Also, to illustrate this point that running Group-LASSO on the $\vec{y}^{\,(i)}_j$ features (Y-GL) is less robust to noise and adaptive to smoothness, we ran noisier trails using the configuration of $(p,N,n)=(1000,500,25)$. We increased the standard deviation of the noise on grid function observation and on the response to be $5$ and $1$ respectively. Under these conditions the FuSSO was able to recover the support in $49\%$ of the trails were as Y-GL recovered the support in $32\%$ of the trails. Furthermore the FuSSO had a $\Delta^{(t)}_\lambda=.0743$ compared to $\Delta^{(t)}_\lambda=.0254$ for Y-GL. 

\subsection{Neurological Data}

\begin{figure}[h]
        \centering
        \subfigure[Example ODF]{\includegraphics[width=.23\textwidth]{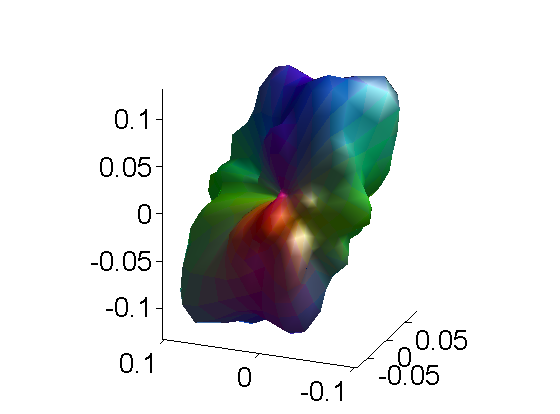}\label{fig:odf}}
        \subfigure[Ages]{\includegraphics[width=.225\textwidth]{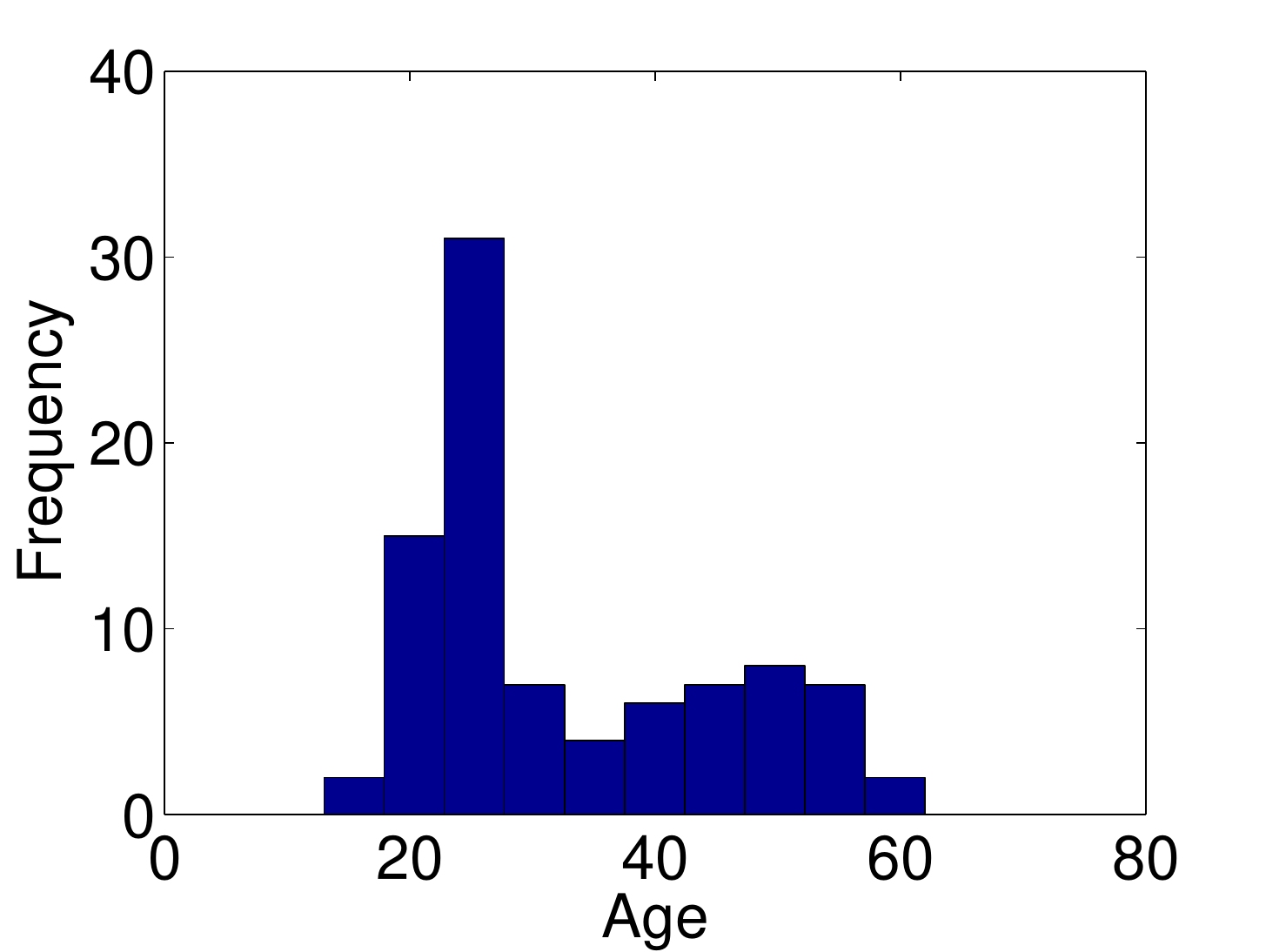}\label{fig:ages}}
        \subfigure[Voxels in support]{\includegraphics[width=.21\textwidth]{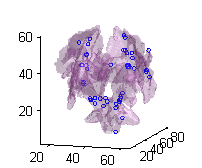}\label{fig:active}}
        \subfigure[Errors]{\includegraphics[width=.225\textwidth]{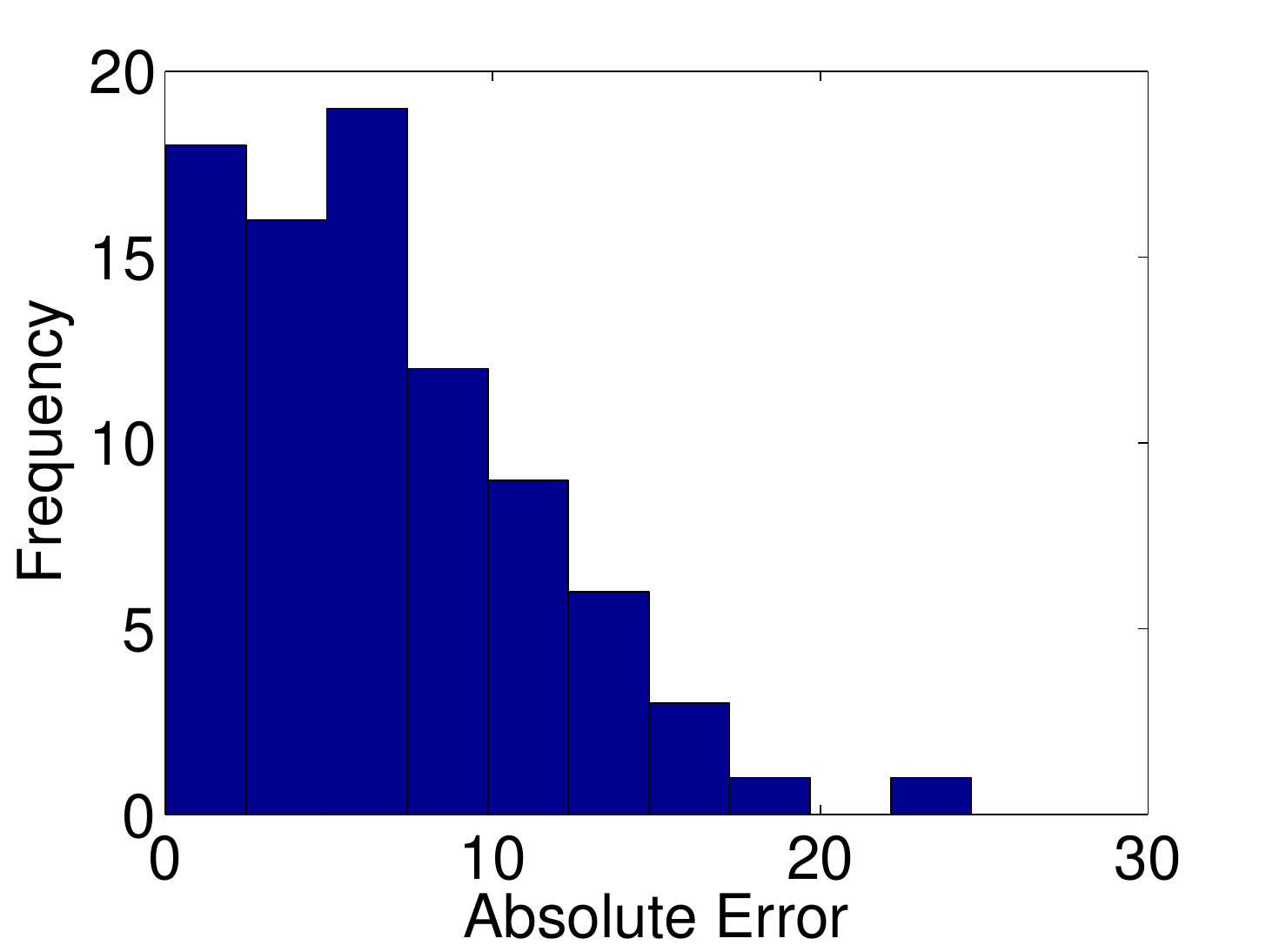}}
        \caption{(a) An example ODF for a voxel. (b) Histogram of ages for subjects. (c) Voxels in the support of model shown in blue. (d) Histogram of held out error magnitudes.}
\vspace{-0.3cm}
\end{figure}

We also tested the FuSSO estimator with a neurological data-set, using a total of 89 subjects \cite{yeh2011ntu}. Subjects ranged in age from 18 to 60 years old (Figure \ref{fig:ages}). Our goal was to learn a regression that maps the dODFs at each white matter voxel for each subject to the subject's age. The dODF is a function represents the amount of water molecules, or spins, undergoing diffusion in different orientations over the $S^2$ sphere\cite{yeh2010generalized}. I.e., each dODF is a function with a $2d$ domain (of azimuth, elevation spherical coordinates) and a range of reals representing the strength of water diffusion at the given orientations (see Figure \ref{fig:odf}). Data was provided for each subject in a template space for white-matter voxels; a total of over 25 thousand voxels' dODFs were regressed on (\ie $p\approx 25000$). We also compared regression using the FuSSO and functional covariates to using the LASSO and real valued covariates. We used the non-functional collection of quantitative anisotropy (QA) values for the same white matter voxels as with dODF functions. QA values are the estimated amount of spins that undergo diffusion in the direction of the principle fiber orientation, i.e., the peak of the dODF; QAs have been used as a measure of white matter integrity in the underlying voxel hence making for a descriptive and effective summary statistic of an dODF function for age regression \cite{yeh2010generalized}.

The projection coefficients for the dODFs at each voxel were estimated using the cosine basis. The FuSSO estimator gave a cross-validated MSE of $70.855$, where the variance for age was $156.4265$; selected voxels in the support may be seen in Figure \ref{fig:active}. The LASSO estimate using QA values gave a cross-validated MSE of $77.1302$. Thus, one may see that considering the entire functional data gave us better results for age regression. We note that we were unable to use the naive approach of Y-GL in this case because of memory constraints and the fact that function evaluation points did not lie on a $2d$ square grid.

\vspace{-0mm}
\section{Conclusion}
In conclusion, this paper presents the FuSSO, a functional analogue to the LASSO. The FuSSO allows one to efficiently find a sparse set of functional input covariates to regress a real-valued response against. The FuSSO makes no parametric assumptions about the nature of input functional covariates and assumes a linear form to the mapping of functional covariates to the response. We provide a statistical backing for use of the FuSSO via proof of asymptotic sparsistency.

\subsubsection*{Acknowledgements}
This work is supported in part by NSF grants IIS1247658 and IIS1250350.
\clearpage

\bibliography{refs}
\bibliographystyle{plain}

\clearpage
\section*{Supplemental Materials}

\subsection*{Lemmata}

\lemma{thm:exp_bnd} Let $X$ be a non-negative r.v. and $\calC$ be an measurable event, then $\EE{X | \calC }\bP(\calC) \leq \EE{X} $. 
\begin{proof}
\begin{align*}
\EE{X} = \EE{X | \calC }\bP(\calC) + \EE{X | \calC^c }\bP(\calC^c) \geq \EE{X | \calC }\bP(\calC)
\end{align*}
\end{proof}

\lemma{thm:sumkrondelt}: $\frac{1}{n}\sum_{k=1}^n \varphi_{m}(k/n)\varphi_{l}(k/n) = \I\{l=m\}$, for $1\leq l,m \leq n-1$. 
\proof{See Lemma 1.7 in \cite{tsybakov2008introduction}.}

\lemma{thm:noisedist}: Let $H_j$ be the $N \times M_n$ matrix with entries $H_{j}(i,m) = \eta^{(i)}_{jm} = \frac{1}{n} \vec{\varphi}_m^T  \xi^{(i)}_j $, then its rows $H_{j}^{(i)} \iid \calN(0,\frac{\sigma_\xi^2}{n}I)$.
\begin{proof}
$H_{j}^{(i)} = \frac{1}{n}[\vec{\varphi}_1 \ldots \vec{\varphi}_{M_n}]^T \xi^{(i)}_j$, hence it is clearly Gaussian with mean $0$. Furthermore,
\begin{align*}
\E[H_{jl}^{(i)}H_{jm}^{(i)}] =& \E[(\tfrac{1}{n}\vec{\varphi}_l^T  \xi^{(i)}_j)(\tfrac{1}{n} \vec{\varphi}_m^T  \xi^{(i)}_j)]\\
=& \tfrac{1}{n}\vec{\varphi}_l^T \E[ (\xi^{(i)}_j)(\xi^{(i)}_j)^T]\tfrac{1}{n} \vec{\varphi}_m \\
=& \tfrac{1}{n^2}\vec{\varphi}_l^T (\sigma_\xi^2I) \vec{\varphi}_m \\
=& \tfrac{\sigma_\xi^2}{n}(\tfrac{1}{n}\vec{\varphi}_l^T \vec{\varphi}_m) \\
=& \tfrac{\sigma_\xi^2}{n} \I\{l=m\},
\end{align*}
where the last line follows from Lemma \ref{thm:sumkrondelt}. Furthermore, $H_{S}^{(i)} \iid \calN(0,\frac{\sigma_\xi^2}{n}I)$ directly follows using Lemma \ref{thm:noisedist} and the fact that $\xi^{(i)}_j$ are independent over $j$ as well as $i$ indices.
\end{proof}

\lemma{thm:bndHmax} $\bP\left(\norm{H}_{\max} \geq n^\sa \right) \leq 2 \sigma_\xi pM_n n^{\sa - \sfrac{1}{2}} e^{-\frac{n^{1-2\sa}}{2\sigma_\xi^2}}$
\begin{proof}
\begin{align*}
\bP\left(\norm{H}_{\max} \geq n^\sa \right) \leq& \bP\left(\cup_{ij} \{ |H_{Sj}^{(i)}| \geq n^{-a} \}\right) 
\\
 \leq& \sum_{ij} \bP\left(  |H_{j}^{(i)}| \geq n^{-a} \right)\\
=& pM_n \bP\left(  \frac{\sigma_\xi}{\sqrt{n}}|Z| \geq n^{-a} \right) \\
\leq& 2 \sigma_\xi pM_n n^{\sa - \sfrac{1}{2}} e^{-\frac{n^{1-2\sa}}{2\sigma_\xi^2}},
\end{align*}
\end{proof}
where $Z\sim\calN(0,1)$, and the last line follows from a Gaussian Tail inequality.

\lemma{thm:biasmax} $\norm{E_j}_{\max}\leq C_Q n^{-\gamma+1/2}$, where $C_Q\in(0,\infty)$ is a constant depending on $Q$. 
\proof{See Lemma 1.8 in \cite{tsybakov2008introduction}.}

\lemma{thm:beta} $\norm{\beta^{*}_S}_2^2 \leq Qs$. 
\begin{proof}
\begin{align*}
\norm{\beta^{*}_S}_2^2 =& \sum_{j\in S}\norm{\beta^{*}_j}_2^2 = \sum_{j\in S} \sum_{m=1}^{M_n} \beta^{*2}_{jm} \leq \sum_{j\in S} \sum_{m=1}^{M_n} c_k^2\beta^{*2}_{jm} \\
\leq& Qs
\end{align*}
\end{proof}

\lemma{thm:noiseeigs}: $\exists N_0,n_0,\tilde{C}_{\min}, \tilde{C}_{\max}$, $ 0<\tilde{C}_{\min}\leq \tilde{C}_{\max} < \infty$, $0<\tdelta\leq1$ s.t. if $\norm{H}_{\max}<n^{-a}$, and $N>N_0$, $n>n_0$ then
\begin{align}
\Lambda_{\max} \left(\tfrac{1}{N}\tilde{A}_S^T\tilde{A}_S\right) \leq \tilde{C}_{\max} <& \infty \\
\Lambda_{\min} \left(\tfrac{1}{N}\tilde{A}_S^T\tilde{A}_S\right) \geq \tilde{C}_{\min} >& 0\\
\forall j \in S^c,\ \Norm{(\tfrac{1}{N}\tilde{A}_{j}^T\tilde{A}_S)(\tfrac{1}{N}\tilde{A}_S^T\tilde{A}_S)^{-1}}_2 \leq&  \frac{1-\tdelta}{\sqrt{s}}
\end{align}
\begin{proof}
First, note that by the Courant-Fischer-Weyl min-max principle (e.g. \cite{bhatia1997matrix}), for symmetric real matrices $B, C$ we have that:
\begin{align*}
\Lambda_{\max}(B+C) =& \max_{\norm{x}=1} x^T(B+C)x \\
=& \max_{\norm{x}=1} x^TBx+x^TCx \\
\leq& \max_{\norm{x}=1} x^TBx+ \max_{\norm{x}=1}x^TCx \\
=& \Lambda_{\max}(B) + \Lambda_{\max}(C)
\end{align*}
and,
\begin{align*}
\Lambda_{\min}(B+C) =& \min_{\norm{x}=1} x^T(B+C)x \\
=& \min_{\norm{x}=1} x^TBx+x^TCx \\
\geq & \min_{\norm{x}=1} x^TBx+ \min_{\norm{x}=1}x^TCx \\
=& \Lambda_{\min}(B) + \Lambda_{\min}(C).
\end{align*}
Thus,
\begin{align}
& \Lambda_{\max} \left(\tfrac{1}{N}\tilde{A}_S^T\tilde{A}_S\right) \nonumber\\
&\leq  \Lambda_{\max}(\tfrac{1}{N}A_S^TA_S) \label{eq:a2}\\
&+\Lambda_{\max}\left(\tfrac{1}{N}((E_S+H_S)^TA_S + A_S^T(E_S+H_S)) \right) \label{eq:aerr} \\
& + \Lambda_{\max}\left(\tfrac{1}{N}(E_S+H_S)^T(E_S+H_S) \right). \label{eq:err2}
\end{align}
Since the term in (\ref{eq:a2}) is bounded by (\ref{eq:maxeig}), we need only show that (\ref{eq:aerr}) and (\ref{eq:err2}) are bounded for large enough $N$, $n$. Similarly, we have that:
\begin{align}
& \Lambda_{\min} \left(\tfrac{1}{N}\tilde{A}_S^T\tilde{A}_S\right) \nonumber\\
&\geq  \Lambda_{\min}(\tfrac{1}{N}A_S^TA_S) \label{eq:mina2}\\
&+ \Lambda_{\min}\left(\tfrac{1}{N}((E_S+H_S)^TA_S + A_S^T(E_S+H_S)) \right) \label{eq:minaerr} \\
& + \Lambda_{\min}\left(\tfrac{1}{N}(E_S+H_S)^T(E_S+H_S) \right). \label{eq:minerr2}
\end{align}
Since the term in (\ref{eq:mina2}) is bounded by (\ref{eq:mineig}) and (\ref{eq:minerr2}) is a positive semi-definite matrices, it suffices to show that 
\begin{align*}
\Lambda_{\min}\left(\tfrac{1}{N}((E_S+H_S)^TA_S + A_S^T(E_S+H_S) )\right) > -C_{\min}
\end{align*}
for large enough $N$, $n$. Note further, that for symmetric matrix $B$: 
\begin{align*}
\Lambda_{\max}(B)\leq& \max_{\norm{x}=1}|x^TBx|\ \mathrm{and} \\
\Lambda_{\min}(B)\geq& -\max_{\norm{x}=1}|x^TBx|.
\end{align*} 
Hence we will use the maximum absolute Rayleigh quotient ($x^TBx$) control bounds on the expected eigenvalues. Note that:
\begin{align*}
&\max_{\norm{x}=1}|x^T((E_S+H_S)^TA_S + A_S^T(E_S+H_S) )x|\leq  \\
& \max_{\norm{x}=1}|((E_S+H_S)x)^T(A_Sx)|+|(A_Sx)^T((E_S+H_S)x)| \\
&=2\max_{\norm{x}=1}|((E_S+H_S)x)^T(A_Sx)| \\
&\leq 2\max_{\norm{x}=1}\norm{(E_S+H_S)x}_2\norm{A_Sx}_2\\
&\leq 2\left(\max_{\norm{x}=1}\norm{A_Sx}_2 \right) \left(\max_{\norm{x}=1}\norm{(E_S+H_S)x}_2\right) \\
&\leq 2\left(\Lambda_{\max}(A_S^TA_S)\right)^{\frac{1}{2}} \left(\sqrt{N}\max_{\norm{x}=1}\norm{(E_S+H_S)x}_\infty\right) \\
&\leq 2\sqrt{NC_{\max}} \left(\sqrt{N}\max_{\norm{x}=1,i}|(E_S^{(i)}+H_S^{(i)})^Tx|\right) \\
&\leq 2N\sqrt{C_{\max}} \left(\max_{\norm{x}=1,i}\norm{E_S^{(i)}+H_S^{(i)}}_2\norm{x}_2\right) \\
&\leq 2N\sqrt{C_{\max}}\sqrt{sM_n}(C_Q n^{-\gamma+1/2} + n^{-\sa}).
\end{align*}
Similarly,
\begin{align*}
&\max_{\norm{x}=1}|x^T(E_S+H_S)^T(E_S+H_S)x|\leq  \\
&sM_nN(C_Q n^{-\gamma+1/2} + n^{-\sa})^2.
\end{align*}

Thus,
\begin{align*}
\Lambda_{\max} \left(\tfrac{1}{N}\tilde{A}_S^T\tilde{A}_S\right) &\leq C_{\max} \\
&+ 2\sqrt{C_{\max}sM_n}(C_Q n^{-\gamma+1/2} + n^{-\sa}) \\
&+ sM_n(C_Q n^{-\gamma+1/2} + n^{-\sa})^2 \\
&\leq \tilde{C}_{\max},
\end{align*}
and
\begin{align*}
\Lambda_{\min} \left(\tfrac{1}{N}\tilde{A}_S^T\tilde{A}_S\right) &\ge C_{\min} \\
&- 2\sqrt{C_{\max}sM_n}(C_Q n^{-\gamma+1/2} + n^{-\sa}) \\
&\ge \tilde{C}_{\min},
\end{align*}
for large enough $n,N$ and appropriate $\tilde{C}_{\max},\tilde{C}_{\min}$ using our assumptions.
Let $\norm{\cdot}=\norm{\cdot}_2$ below. Hence,
\begin{align*}
&\Norm{(\tfrac{1}{N}\tilde{A}_j^T\tilde{A}_S)(\tfrac{1}{N}\tilde{A}_S^T\tilde{A}_S)^{-1}} \\ &=\Norm{(\tfrac{1}{N}\tilde{A}_j^T\tilde{A}_S)(\tfrac{1}{N}A_S^TA_S)^{-1}(\tfrac{1}{N}A_S^TA_S)(\tfrac{1}{N}\tilde{A}_S^T\tilde{A}_S)^{-1}} \\
&\leq \Norm{(\tfrac{1}{N}\tilde{A}_j^T\tilde{A}_S)(\tfrac{1}{N}A_S^TA_S)^{-1}}\Norm{(\tfrac{1}{N}A_S^TA_S)(\tfrac{1}{N}\tilde{A}_S^T\tilde{A}_S)^{-1}}.
\end{align*}
Also,
\begin{align}
&\Norm{(\tfrac{1}{N}\tilde{A}_j^T\tilde{A}_S)(\tfrac{1}{N}A_S^TA_S)^{-1}} \nonumber\\
&= \Norm{\tfrac{1}{N}(A_j+E_j+H_j)^T(A_S+E_S+H_S)(\tfrac{1}{N}A_S^TA_S)^{-1}} \nonumber\\
&\leq \Norm{\tfrac{1}{N}A_j^TA_S(\tfrac{1}{N}A_S^TA_S)^{-1}} \nonumber\\
&+\Norm{\tfrac{1}{N}A_j^T(E_S+H_S)(\tfrac{1}{N}A_S^TA_S)^{-1}} \nonumber\\
&+\Norm{\tfrac{1}{N}(E_j+H_j)^TA_S(\tfrac{1}{N}A_S^TA_S)^{-1}} \nonumber\\
&+\Norm{\tfrac{1}{N}(E_j+H_j)^T(E_S+H_S)(\tfrac{1}{N}A_S^TA_S)^{-1}} \nonumber\\
&\leq\frac{1-\delta}{\sqrt{s}}+ \Norm{\tfrac{1}{\sqrt{N}}A_j^T}\Norm{\tfrac{1}{\sqrt{N}}(E_S+H_S)}\Norm{(\tfrac{1}{N}A_S^TA_S)^{-1}} \nonumber\\
&+ \Norm{\tfrac{1}{\sqrt{N}}(E_j+H_j)}\Norm{\tfrac{1}{\sqrt{N}}A_S}\Norm{(\tfrac{1}{N}A_S^TA_S)^{-1}} \nonumber\\
&+ \Norm{\tfrac{1}{N}(E_j+H_j)^T(E_S+H_S)}\Norm{(\tfrac{1}{N}A_S^TA_S)^{-1}} \nonumber\\
&\leq\frac{1-\delta}{\sqrt{s}}+ \frac{\sqrt{C_{\max}}}{C_{\min}}\sqrt{sM_n}(C_Q n^{-\gamma+1/2} + n^{-\sa}) \label{eq:terms1}\\
&+ \frac{\sqrt{C_{\max}}}{C_{\min}}\sqrt{M_n}(C_Q n^{-\gamma+1/2} + n^{-\sa}) \\
&+ \frac{1}{C_{\min}}\sqrt{s}M_n(C_Q n^{-\gamma+1/2} + n^{-\sa})^2. 
\end{align}
Also,
\begin{align*}
&\tfrac{1}{N}A_S^TA_S \\
&=\tfrac{1}{N}\tilde{A}_S^T\tilde{A}_S-\tfrac{1}{N}A_S^T(E_S+H_S)\\
&-\tfrac{1}{N}(E_S+H_S)^TA_S-\tfrac{1}{N}(E_S+H_S)^T(E_S+H_S).
\end{align*}
Thus,
\begin{align}
&\Norm{(\tfrac{1}{N}A_S^TA_S)(\tfrac{1}{N}\tilde{A}_S^T\tilde{A}_S)^{-1}} \nonumber\\
&\leq 1 + \Norm{\tfrac{1}{N}A_S^T(E_S+H_S)(\tfrac{1}{N}\tilde{A}_S^T\tilde{A}_S)^{-1}} \nonumber\\
&+\Norm{\tfrac{1}{N}(E_S+H_S)^TA_S(\tfrac{1}{N}\tilde{A}_S^T\tilde{A}_S)^{-1}} \nonumber\\
&+\Norm{\tfrac{1}{N}(E_S+H_S)^T(E_S+H_S)(\tfrac{1}{N}\tilde{A}_S^T\tilde{A}_S)^{-1}} \nonumber\\
&\leq 1 + \frac{\sqrt{C_{\max}}}{C_{\min}}\sqrt{sM_n}(C_Q n^{-\gamma+1/2} + n^{-\sa})\\
&+ \frac{\sqrt{C_{\max}}}{C_{\min}}\sqrt{sM_n}(C_Q n^{-\gamma+1/2} + n^{-\sa})\\
&+ \frac{1}{C_{\min}}sM_n(C_Q n^{-\gamma+1/2} + n^{-\sa})^2 \label{eq:terms2}.
\end{align}
By our assumptions all terms in \eqref{eq:terms1}-\eqref{eq:terms2}, except $\frac{1-\delta}{\sqrt{s}}$, are going to zero. Hence, keeping leading terms, one may see that
\begin{align*}
&\Norm{(\tfrac{1}{N}\tilde{A}_j^T\tilde{A}_S)(\tfrac{1}{N}\tilde{A}_S^T\tilde{A}_S)^{-1}} \\
&\leq \frac{1-\delta}{\sqrt{s}} + O\left(\sqrt{sM_n}(n^{-\gamma+1/2} + n^{-\sa})\right)\\
&+ O\left(\sqrt{s}M_n( n^{-\gamma+1/2} + n^{-\sa})^2\right) \\
&\leq \frac{1-\tdelta}{\sqrt{s}}
\end{align*}
for large enough $n$ and appropriate $\tdelta$.
\end{proof}

\lemma{thm:Vterm} $\E\left[\norm{\SigSS(\tfrac{1}{N}\tilde{A}_S^T)V}_\infty \Given \calB \right]\bP(\calB) = O\left( \frac{s^{\sfrac{3}{2}}}{M_n^{2\gamma-\sfrac{1}{2}}} \right)$ .
\begin{proof}
First note that:
\begin{align*}
\norm{\SigSS(\tfrac{1}{N}\tilde{A}_S^T)V}_\infty  \leq \norm{\SigSS}_\infty \norm{(\tfrac{1}{N}\tilde{A}_S^T)}_\infty\norm{V}_\infty.
\end{align*}
We have that
\begin{align*}
|V_i| &= \left| \sum_{j\in S} \sum_{m=M_n+1}^{\infty} \alpha_{jm}^{(i)}\beta_{jm}^{*} \right| \leq  \sum_{j\in S} \sum_{m=M_n+1}^{\infty} \left|\alpha_{jm}^{(i)}\beta_{jm}^{*} \right| \\
&\leq  \sum_{j\in S} \left(\sum_{m=M_n+1}^{\infty} \alpha_{jm}^{(i)2} \right)^{\frac{1}{2}}  \left(\sum_{m=M_n+1}^{\infty} \beta_{jm}^{*2}\right)^{\frac{1}{2}},
\end{align*}
and
\begin{align*}
&\frac{1}{M_n^{2\gamma}}\left(\sum_{m=M_n+1}^{\infty} M_n^{2\gamma} \alpha_{jm}^{(i)2} \right)^{\frac{1}{2}}  \left(\sum_{m=M_n+1}^{\infty} M_n^{2\gamma} \beta_{jm}^{*2}\right)^{\frac{1}{2}} \\
&\leq \frac{1}{M_n^{2\gamma}}\left(\sum_{m=1}^{\infty} c_k^2 \alpha_{jm}^{(i)2} \right)^{\frac{1}{2}}  \left(\sum_{m=1}^{\infty} c_k^2 \beta_{jm}^{*2}\right)^{\frac{1}{2}}\\
&\leq \frac{Q}{M_n^{2\gamma}}.
\end{align*}
Thus $|V_i|  \leq \frac{Qs}{M_n^{2\gamma}}$. 
Also,
\begin{align*}
\norm{(\tfrac{1}{N}\tilde{A}_S^T)}_\infty &\leq \frac{1}{N}(\norm{A_S^T}_\infty+\norm{E_S^T}_\infty+\norm{H_S^T}_\infty)\\
&\leq Q + C_Q n^{-\gamma+1/2}+n^{-\sa}.
\end{align*}
Hence,
\begin{align}
&\E\left[\norm{\SigSS(\tfrac{1}{N}\tilde{A}_S^T)V}_\infty \Given \calB \right] \nonumber \\
&\leq \frac{\sqrt{sM_n} }{\tilde{C}_{\min}}\left(Q + C_Q n^{-\gamma+1/2}+n^{-\sa} \right)\frac{Qs}{M_n^{2\gamma}}.
\end{align}
\end{proof}

\lemma{thm:epsterm}  $\EE{\norm{\SigSS(\tfrac{1}{N}\tilde{A}_S^T)\epsilon}_\infty \Given \calB}\bP(\calB)  = O\left(\sqrt{\log(sM_n)/N}\right)$.
\begin{proof}
Note that given $H_S$,  $Z = \SigSS(\tfrac{1}{N}\tilde{A}_S^T)\epsilon$ is normal with mean 0 and co-variance matrix:
\begin{align*}
&\EE{\SigSS(\tfrac{1}{N}\tilde{A}_S^T)\epsilon\epsilon^T(\tfrac{1}{N}\tilde{A}_S)\SigSS | H_S} \\
&= \SigSS(\tfrac{1}{N}\tilde{A}_S^T)(\sigma_\epsilon^2I)(\tfrac{1}{N}\tilde{A}_S)\SigSS \\
&= \frac{\sigma_\epsilon^2}{N}\SigSS(\tfrac{1}{N}\tilde{A}_S^T\tilde{A}_S)\SigSS \\
&= \frac{\sigma_\epsilon^2}{N}\SigSS.
\end{align*}
Hence, given $H_S$ and $\calB$
\begin{align*}
\max_{i}\Var\left[Z_i \right] =& \max_{i}\frac{\sigma_\epsilon^2}{N}e_i^T\SigSS e_i \leq \frac{\sigma_\epsilon^2}{N} \left(\Lambda_{\min}(\tilde{\Sigma}_{SS})\right)^{-1}\\
&\leq \frac{\sigma_\epsilon^2}{N} \tilde{C}_{\min}^{-1}.
\end{align*}
And so \cite{ledoux1991probability},
\begin{align*}
&\EE{ \norm{Z}_\infty \Given \calB}\\
&= \EE{ \EE{\norm{Z}_\infty \Given \calB, H_S} \Given \calB} \\
&\leq \EE{  \EE{ 3\sqrt{\log(sM_n)\norm{\Var[Z]}_\infty}\Given \calB,H_S} \Given \calB} \\
&\leq \EE{  3\sigma_\epsilon \sqrt{\frac{\log(sM_n)}{N\tilde{C}_{\min}}} \Given \calB} \\
&=3\sigma_\epsilon \sqrt{\frac{\log(sM_n)}{N\tilde{C}_{\min}}}.
\end{align*}
\end{proof}


\lemma{thm:uH} $\bP\left(\max_{j\in S^c} \norm{\mu_j^H }_2 \geq 1-\frac{\tdelta}{2} \right) \rightarrow 0 $
\begin{proof}
\begin{align*}
&\bP\left(\max_{j\in S^c} \norm{\mu_j^H }_2 \geq 1-\frac{\tdelta}{2} \right) \\
&= \bP\left(\max_{j\in S^c} \norm{\mu_j^H }_2-(1-\tdelta) \geq \frac{\tdelta}{2} \right) \\
&\leq \bP\left(\max_{j\in S^c} \norm{\mu_j^H }_2-(1-\tdelta) \geq \frac{\tdelta}{2} \Given \calB \right)\bP\left(\calB\right)+\bP\left(\calB^c\right)\\
&\leq \frac{2}{\tdelta} \EE{\max_{j\in S^c} \norm{\mu_j^H }_2-(1-\tdelta) \Given \calB }\bP\left(\calB\right)+\bP\left(\calB^c\right),
\end{align*}
Recall that
\begin{align*}
\hat{u}_j =&\tfrac{1}{\Nlambda}\SigjS\SigSS\Big( \tfrac{1}{N}\tilde{A}_S^T(E_S+H_S)\beta^{*}_S -\tfrac{1}{N}\tilde{A}_S^TV\\ &-\tfrac{1}{N}\tilde{A}_S^T\epsilon+\Nlambda\hat{u}_S\Big)-\tfrac{1}{\Nlambda N}\tilde{A}_j^T(E_S+H_S)\beta^{*}_S \\
&+ \tfrac{1}{\Nlambda N} \tilde{A}_j^T (V+\epsilon),
\end{align*}
where $\SigjS=\tfrac{1}{N}\tilde{A}_j^T\tilde{A}_S$.
Let $\mu_j^H \equiv \EE{\hat{u}_j \given H}=$
\begin{align*}
&\SigjS\SigSS \left( \tfrac{1}{\Nlambda N}\tilde{A}_S^T(E_S+H_S)\beta^{*}_S -\tfrac{1}{\Nlambda N}\tilde{A}_S^TV+\hat{u}_S\right) \\
&-\tfrac{1}{\Nlambda N}\tilde{A}_j^T(E_S+H_S)\beta^{*}_S + \tfrac{1}{\Nlambda N} \tilde{A}_j^T V.
\end{align*}
Note that
\begin{align*}
\norm{\mu_j^H}_2 \leq& \norm{\SigjS\SigSS}_2 \Big( \norm{\tfrac{1}{\Nlambda N}\tilde{A}_S^T(E_S+H_S)\beta^{*}_S}_2\\
&+\norm{\tfrac{1}{\Nlambda N}\tilde{A}_S^TV}_2 + \norm{\hat{u}_S}_2\Big)\\
&+\norm{\tfrac{1}{\Nlambda N}\tilde{A}_j^T(E_S+H_S)\beta^{*}_S}_2 + \norm{\tfrac{1}{\Nlambda N} \tilde{A}_j^T V}_2.
\end{align*}
Given $\calB$,
\begin{align*}
\norm{\mu_j^H }_2 \leq& \frac{1-\tdelta}{\sqrt{s}}\Big( \norm{\tfrac{1}{\Nlambda N}\tilde{A}_S^T(E_S+H_S)\beta^{*}_S}_2\\
&+\norm{\tfrac{1}{\Nlambda N}\tilde{A}_S^TV}_2 + \sqrt{s}\Big)\\
&+\norm{\tfrac{1}{\Nlambda N}\tilde{A}_j^T(E_S-H_S)\beta^{*}_S}_2 + \norm{\tfrac{1}{\Nlambda N} \tilde{A}_j^T V}_2\\
=& 1-\tdelta + \tfrac{1-\tdelta}{\sqrt{s}}\norm{\tfrac{1}{\Nlambda N}\tilde{A}_S^T(E_S+H_S)\beta^{*}_S}_2 \\
&+\tfrac{1-\tdelta}{\sqrt{s}}\norm{\tfrac{1}{\Nlambda N}\tilde{A}_S^TV}_2\\
&+\norm{\tfrac{1}{\Nlambda N}\tilde{A}_j^T(E_S+H_S)\beta^{*}_S}_2 + \norm{\tfrac{1}{\Nlambda N} \tilde{A}_j^T V}_2
\end{align*}
and so:
\begin{align*}
&\EE{\max_{j\in S^c} \norm{\mu_j^H }_2-(1-\tdelta) \Given \calB} \bP(\calB)\\
&\leq \tfrac{1-\tdelta}{\sqrt{s}}\EE{\norm{\tfrac{1}{\Nlambda N}\tilde{A}_S^T(E_S+H_S)\beta^{*}_S}_2\Given \calB} \bP(\calB)\\
&+\tfrac{1-\tdelta}{\sqrt{s}}\EE{\norm{\tfrac{1}{\Nlambda N}\tilde{A}_S^TV}_2\Given \calB}\bP(\calB)\\
&+\EE{\max_{j\in S^c}\norm{\tfrac{1}{\Nlambda N}\tilde{A}_j^T(E_S+H_S)\beta^{*}_S}_2 \Given \calB}\bP(\calB)\\
&+ \EE{\max_{j\in S^c}\norm{\tfrac{1}{\Nlambda N} \tilde{A}_j^T V}_2 \Given \calB}\bP(\calB).
\end{align*}
First, note that
\begin{align*}
&\EE{\norm{\tfrac{1}{\Nlambda N}A_S^T(E_S+H_S)\beta^{*}_S}_2} \bP(\calB)\\
&\leq  \tfrac{1}{\Nlambda \sqrt{N}} \EE{ \norm{\tfrac{1}{\sqrt{N}}A_S^T}_2 \norm{(E_S+H_S)\beta^{*}_S}_2 } \bP(\calB)\\
&\leq \tfrac{\sqrt{C_{max}}}{\Nlambda \sqrt{N}} \EE{ \sqrt{N}\norm{(E_S+H_S)\beta^{*}_S}_\infty} \bP(\calB) \\
&= O\left( \frac{1}{\Nlambda} \left(s\sqrt{M_n} n^{-\gamma +1/2} +   \sqrt{\frac{s\log(N)}{n}}\right) \right).
\end{align*}
Moreover,
\begin{align*}
&\EE{\norm{\tfrac{1}{\Nlambda N}(E_S+H_S)^T(E_S+H_S)\beta^{*}_S}_2 \Given \calB}\bP(\calB) \\
&\leq \tfrac{\bP(\calB)}{\Nlambda N} \EE{ \sqrt{sM_n}\norm{(E_S+H_S)^T(E_S+H_S)\beta^{*}_S}_\infty \Given \calB} \\
&= O\Bigg( \frac{\sqrt{sM_n}}{\Nlambda } \left(  s\sqrt{M_n} n^{-\gamma +1/2} + \sqrt{\frac{s\log(N)}{n} } \right)\\
&\quad \left(n^{-\gamma+1/2}+n^{-\sa}\right)\Bigg).
\end{align*}
Thus,
\begin{align*}
&\tfrac{1-\tdelta}{\sqrt{s}}\EE{\norm{\tfrac{1}{\Nlambda N}\tilde{A}_S^T(E_S+H_S)\beta^{*}_S}_2\Given \calB} \bP(B) \\
&=O\left( \frac{1}{\Nlambda}\sqrt{sM_n} n^{-\gamma +1/2} +   \frac{1}{\Nlambda}\sqrt{\frac{\log(N)}{n}}\right)\\
&+O \left(  \frac{s M_n}{\Nlambda}  n^{-2\gamma +1} + \frac{\sqrt{sM_n\log(N) }}{\Nlambda n^{\gamma}} \right)\\
&+O \left(  \frac{s M_n}{\Nlambda n^{\gamma+\sa -\sfrac{1}{2}}}   + \frac{\sqrt{sM_n\log(N)}}{\Nlambda n^{\sa +\sfrac{1}{2}}} \right).
\end{align*}
Similarly,
\begin{align*}
&\EE{\norm{\tfrac{1}{\Nlambda N}A_j^T(E_S+H_S)\beta^{*}_S}_2\Given \calB} \bP(\calB)\\
&\leq \frac{\sqrt{N}}{\Nlambda \sqrt{N}}\norm{\tfrac{1}{\sqrt{N}}A_j^T}_2\EE{\norm{(E_S+H_S)\beta^{*}_S}_\infty\Given \calB} \bP(\calB)\\
&= O\left( \frac{1}{\Nlambda} \left(  s\sqrt{M_n} n^{-\gamma +1/2} + \sqrt{\frac{s\log(N)}{n} } \right) \right),
\end{align*}
and,
\begin{align*}
&\EE{\norm{\tfrac{1}{\Nlambda N}(E_j+H_j)^T(E_S+H_S)\beta^{*}_S}_2\Given \calB} \bP(\calB)\\
&\leq \frac{\sqrt{M_n}}{\Nlambda N}\EE{\norm{(E_j+H_j)^T(E_S+H_S)\beta^{*}_S}_\infty\Given \calB} \bP(\calB)\\
&= O\Bigg( \frac{\sqrt{M_n}}{\Nlambda } \left(  s\sqrt{M_n} n^{-\gamma +1/2} + \sqrt{\frac{s\log(N)}{n} } \right)\\
&\quad \left(n^{-\gamma+1/2}+n^{-\sa}\right)\Bigg).
\end{align*}
Hence,
\begin{align*}
&\EE{\norm{\tfrac{1}{\Nlambda N}\tilde{A}_j^T(E_S+H_S)\beta^{*}_S}_2\Given \calB} \bP(B) \\
&=O\left(\frac{s\sqrt{M_n}}{\Nlambda } n^{-\gamma +1/2} +   \frac{1}{\Nlambda} \sqrt{\frac{s\log(N)}{n}}\right)\\
&+O \left(  \frac{s M_n}{\Nlambda}  n^{-2\gamma +1} + \frac{\sqrt{sM_n\log(N) }}{\Nlambda n^{\gamma}} \right)\\
&+O \left(  \frac{s M_n}{\Nlambda n^{\gamma+\sa -\sfrac{1}{2}}}   + \frac{\sqrt{sM_n\log(N)}}{\Nlambda n^{\sa +\sfrac{1}{2}}} \right)
\end{align*}
Also, $\EE{\norm{\tfrac{1}{\Nlambda N}\tilde{A}_S^TV}_2 \Given \calB} \bP(\calB) $
\begin{align*}
&\leq \frac{\sqrt{sM_n}}{\Nlambda}\EE{\norm{\tfrac{1}{N}\tilde{A}_S^TV}_\infty \Given \calB} \bP(\calB)\\
&= O\left(\frac{\sqrt{sM_n}}{\Nlambda} \left(Q + C_Q n^{-\gamma+1/2}+n^{-\sa} \right)\frac{Qs}{M_n^{2\gamma}}\right)\\
&= O\left( \frac{s^{\sfrac{3}{2}}}{\Nlambda M_n^{2\gamma-\sfrac{1}{2}}} \left(1+ n^{-\gamma+1/2}+n^{-\sa} \right) \right)\\
&= O\left( \frac{s^{\sfrac{3}{2}}}{\Nlambda M_n^{2\gamma-\sfrac{1}{2}}} \right).
\end{align*}
So, 
\begin{align*}
\tfrac{1-\tdelta}{\sqrt{s}}\EE{\norm{\tfrac{1}{\Nlambda N}\tilde{A}_S^TV}_2 \Given \calB} \bP(\calB) = O\left( \frac{s}{\Nlambda M_n^{2\gamma-\sfrac{1}{2}}} \right).
\end{align*}
Similarly, $\EE{\norm{\tfrac{1}{\Nlambda N}\tilde{A}_j^TV}_2 \Given \calB} \bP(\calB)$
\begin{align*}
&\leq \frac{\sqrt{M_n}}{\Nlambda}\EE{\norm{\tfrac{1}{N}\tilde{A}_j^TV}_\infty \Given \calB} \bP(\calB)\\
&= O\left( \frac{s}{\Nlambda M_n^{2\gamma-\sfrac{1}{2}}} \right).
\end{align*}
Thus,
\begin{align*}
&\frac{2}{\tdelta}\EE{\max_{j\in S^c} \norm{\mu_j^H }_2-(1-\tdelta) \Given \calB} \bP(\calB)\\
&=O\left(\frac{s\sqrt{M_n}}{\Nlambda } n^{-\gamma +1/2} +   \frac{1}{\Nlambda} \sqrt{\frac{s\log(N)}{n}}\right)\\
&+O \left(  \frac{s M_n}{\Nlambda n^{\gamma+\sa -\sfrac{1}{2}}}   + \frac{\sqrt{sM_n\log(N)}}{\Nlambda n^{\sa +\sfrac{1}{2}}} \right)\\
&+O\left( \frac{s}{\Nlambda M_n^{2\gamma-\sfrac{1}{2}}} \right),
\end{align*}
where we used that fact that $\gamma\geq1$. Thus, with assumptions (\ref{eq:bndH_S})-(\ref{eq:l-ass}), $\bP\left(\max_{j\in S^c} \norm{\mu_j^H }_2 \geq 1-\frac{\tdelta}{2} \right) \rightarrow 0$ 
\end{proof}

\lemma{thm:uj} $\bP\left(\max_{j\in S^c} \norm{\hat{u}_j - \mu_j^H }_\infty \geq \frac{\tdelta}{2\sqrt{M_n}} \right) \rightarrow 0$
Note that
\begin{align*}
&\bP\left(\max_{j\in S^c} \norm{\hat{u}_j - \mu_j^H }_\infty \geq \frac{\tdelta}{2\sqrt{M_n}} \right) \\
&\leq \bP\left(\max_{j\in S^c} \norm{\hat{u}_j - \mu_j^H }_\infty \geq \frac{\tdelta}{2\sqrt{M_n}} \Given \calB \right)\bP\left( \calB \right) + \bP\left( \calB^c \right),
\end{align*}
and
\begin{align*}
&\frac{2\sqrt{M_n}}{\tdelta} \EE{\max_{j\in S^c} \norm{\hat{u}_j - \mu_j^H }_\infty  \Given \calB}\\
&=\frac{2\sqrt{M_n}}{\tdelta} \EE{\EE{\max_{j\in S^c} \norm{\hat{u}_j - \mu_j^H }_\infty  \Given \calB, H} \Given \calB}.
\end{align*}
Let 
\begin{align*}
Z_j \equiv& \Nlambda (\hat{u}_j - \mu_j^H)  \\
=&  \tilde{A}_j^T (I - \tilde{A}_S(\tilde{A}_S^T\tilde{A}_S)^{-1}\tilde{A}_S^T )\frac{\epsilon}{N}.
\end{align*}
Thus, given $H$, $Z_j$ is a zero mean Gaussian random variable. Furthermore, given $\calB$, $\max_{k}\Var[Z_{jk}]\leq \sigma_\epsilon^2/N$.

\begin{align*}
\EE{Z_j^TZ_j} =& \frac{1}{N^2}\tilde{A}_j^T(I-\tilde{A}_S(\tilde{A}_S^T\tilde{A}_S)^{-1}\tilde{A}_S^T)\EE{\epsilon \epsilon^T}\\
&(I-\tilde{A}_S(\tilde{A}_S^T\tilde{A}_S)^{-1}\tilde{A}_S^T)\tilde{A}_j \\
=& \frac{\sigma^2}{N^2}\tilde{A}_j^T(I-\tilde{A}_S(\tilde{A}_S^T\tilde{A}_S)^{-1}\tilde{A}_S^T)\\
&(I-\tilde{A}_S(\tilde{A}_S^T\tilde{A}_S)^{-1}\tilde{A}_S^T)\tilde{A}_j \\
=& \frac{\sigma^2}{N^2}\tilde{A}_j^T\Big(I-\tilde{A}_S(\tilde{A}_S^T\tilde{A}_S)^{-1}\tilde{A}_S^T \\
&-\tilde{A}_S(\tilde{A}_S^T\tilde{A}_S)^{-1}\tilde{A}_S^T \\
&+ \tilde{A}_S(\tilde{A}_S^T\tilde{A}_S)^{-1}\tilde{A}_S^T\tilde{A}_S(\tilde{A}_S^T\tilde{A}_S)^{-1}\tilde{A}_S^T \Big)\tilde{A}_j \\
=& \frac{\sigma^2}{N^2}\tilde{A}_j^T\Big(I-\tilde{A}_S(\tilde{A}_S^T\tilde{A}_S)^{-1}\tilde{A}_S^T \\
&-\tilde{A}_S(\tilde{A}_S^T\tilde{A}_S)^{-1}\tilde{A}_S^T + \tilde{A}_S(\tilde{A}_S^T\tilde{A}_S)^{-1}\tilde{A}_S^T \Big)\tilde{A}_j \\
=& \frac{\sigma^2}{N^2}\tilde{A}_j^T\left(I-\tilde{A}_S(\tilde{A}_S^T\tilde{A}_S)^{-1}\tilde{A}_S^T \right)\tilde{A}_j\\
=& \frac{\sigma^2}{N}\left(\frac{1}{N}\tilde{A}_j^T\tilde{A}_j-\frac{1}{N}\tilde{A}_j^T\tilde{A}_S(\tilde{A}_S^T\tilde{A}_S)^{-1}\tilde{A}_S^T\tilde{A}_j \right).
\end{align*}
So, given $\calB$
\begin{align*}
\Var[Z_{jk}] =& \frac{\sigma^2}{N}\left(\frac{1}{N}e_k^T\tilde{A}_j^T\tilde{A}_je_k\right)\\
&-\frac{\sigma^2}{N^2}e_k^T\tilde{A}_j^T\tilde{A}_S(\tilde{A}_S^T\tilde{A}_S)^{-1}\tilde{A}_S^T\tilde{A}_j e_k\\
=& O\left(\frac{1}{N} \right),
\end{align*}
where the last line follows from the fact that $\tilde{A}_j^T\tilde{A}_S(\tilde{A}_S^T\tilde{A}_S)^{-1}\tilde{A}_S^T\tilde{A}_j$ is PSD and $\norm{\frac{1}{N}\tilde{A}_j^T\tilde{A}_j}_{\max}\leq (Q+C_Qn^{-\gamma+\sfrac{1}{2}}+n^{-\sa})^2$.

Hence,
\begin{align*}
\frac{1}{\Nlambda}\EE{\max_{j\in S^c} \norm{Z_j }_\infty  \Given \calB, H} = O\left(\frac{1}{\Nlambda} \sqrt{ \frac{\log((p-s)M_n)}{N}} \right).
\end{align*}
Hence,
\begin{align*}
&\frac{2\sqrt{M_n}}{\tdelta}\E\left[\max_{j\in S^c} \norm{\hat{u}_j - \mu_j^H }_\infty   \right] \\
\leq& O\left(\frac{1}{\Nlambda} \sqrt{ M_n\frac{\log((p-s)M_n)}{N}} \right),
\end{align*}
and so $\bP\left(\max_{j\in S^c} \norm{\hat{u}_j - \mu_j^H }_\infty \geq \frac{\tdelta}{2\sqrt{M_n}} \right) \rightarrow 0$

\end{document}